\documentclass[preprint,12pt]{elsarticle}
\pdfoutput=1

\usepackage[unicode]{hyperref}
\usepackage{amssymb}
\usepackage{amsmath}
\usepackage{graphicx}
\usepackage{enumerate}
\usepackage{url} 
\usepackage{hyperref}
\usepackage[figuresright]{rotating}
\usepackage{graphicx}
\usepackage{enumerate}
\usepackage{natbib}
\usepackage{url} 
\usepackage{hyperref}
\usepackage[title]{appendix}%
\usepackage[figuresright]{rotating}
\usepackage{color}
\usepackage{dsfont}
\usepackage{amsthm}
\usepackage{multirow}
\usepackage{caption}
\usepackage{float}
\usepackage{subcaption}
\usepackage{threeparttable}
\usepackage{utfsym}
\newtheorem{theorem}{Theorem}
\newtheorem{lemma}{Lemma}
\newtheorem{condition}{Condition}
\newtheorem{assumption}{Assumption}
\newtheorem{corollary}{Corollary}
\newtheorem{definition}{Definition}
\RequirePackage{amsthm,amsfonts,amssymb}
\usepackage{balance}
\usepackage{picins}

\journal{Neurocomputing}

\begin{document}

\begin{frontmatter}

\title{Generalization and Risk Bounds for Recurrent Neural Networks}

\author[label1,label2]{Xuewei Cheng}
\ead{xwcheng@hunnu.edu.cn}

\author[label3,label4]{Ke Huang}
\ead{khuan049@ucr.edu}

\author[label3]{Shujie Ma\corref{mycorrespondingauthor}}
\cortext[mycorrespondingauthor]{Corresponding author}
\ead{shujie.ma@ucr.edu}

\address[label1]{MOE-LCSM, School of Mathematics and Statistics, Hunan Normal University, Changsha, China.}

\address[label2]{Key Laboratory of Applied Statistics and Data Science, College of Hunan Province, Hunan Normal University, Changsha, China.}

\address[label3]{Department of Statistics, University of California, Riverside, CA, U.S.A.}

\address[label4]{Neuroscience, Brain and Mind Research Institute, Weill Cornell Medicine, Cornell University, Ithaca, NY, U.S.A.}


\begin{abstract}
Recurrent Neural Networks (RNNs) have achieved great success in the prediction of sequential data. However, their theoretical studies are still lagging behind because of their complex interconnected structures. In this paper, we establish a new generalization error bound for vanilla RNNs, and provide a unified framework to calculate the Rademacher complexity that can be applied to a variety of loss functions. When the ramp loss is used, we show that our bound is tighter than the existing bounds based on the same assumptions on the Frobenius and spectral norms of the weight matrices and a few mild conditions. Our numerical results show that our new generalization bound is the tightest among all existing bounds in three public datasets. Our bound improves the second tightest one by an average percentage of 13.80\% and 3.01\% when the $\tanh$ and ReLU activation functions are used, respectively. Moreover, we derive a sharp estimation error bound for RNN-based estimators obtained through empirical risk minimization (ERM) in multi-class classification problems when the loss function satisfies a Bernstein condition. 
\end{abstract}


\end{frontmatter}

\section{Introduction}\label{S0}
Recurrent Neural Networks (RNNs) are a class of artificial neural networks, which are designed to process time series data and other sequence data. They have been successfully applied in speech recognition \citep{sak2014long,graves2013speech}, natural language processing \citep{sutskever2014sequence,wu2016google}, computer vision \citep{xu2015show,karpathy2015deep}, genome sequencing and biomedicine \citep{cao2018deep,ChenBatsakis2023}, among others. The sequential modeling nature of RNNs makes them significantly different from feed-forward neural networks (FNNs), although both of them have neurons as the basic components. RNNs exploit the internal state (also known as the hidden unit) to process the sequence of inputs, which naturally capture the dependence of the sequence. \cite{allen2019convergence} have proven that RNNs trained with stochastic gradient descent (SGD) can achieve a small training loss when the number of neurons is polynomial in the number of observations of the training data and the maximum of the sequence length.

Although RNNs have achieved great success in many artificial intelligence applications, their theoretical studies are still significantly lagging behind. In this paper, we focus on the investigation of the generalization error bound of vanilla RNNs and the estimation error bound for RNN-based estimators obtained through empirical risk minimization in multi-class classification problems -- the study of both properties is essential to evaluate the predictive ability of a learning algorithm.

In general, the generalization error bound, measuring the difference between expected and empirical risks \citep{vapnik1998statistical,mohri2018}, provides guarantees on the performance of learning algorithms and insights on how well a model trained on finite samples is expected to perform on unseen data. It is a fundamental concept in statistical learning theory \citep{valiant1984theory}. It has been shown in \citep{mohri2018} that the generalization error can be upper bounded by the complexity of the learning model and a stochastic error term, so studying the model complexity is crucial for deriving the generalization error bound. It is worth noting that the investigation of the model complexity of RNNs can be traced back to the late '90s. \cite{dasgupta1995sample} and \cite{koiran1998vapnik}  were early studies that assessed the model complexity of RNNs with simplified structures by the Vapnik-Chervonenkis (VC) dimension \citep{Vapnik71}. The VC dimension provides a measure of model complexity, but it is often restricted to binary functions.  

In recent years, researchers have become more interested in deriving the bounds based on the Rademacher complexity of learning models \citep{scott2014rademacher}, which is data distribution dependent and can provide a tighter bound than the VC dimension (distribution independent) on the generalization error \citep{Huetal2021}.  The development of the bounds started from a relatively easier case: FNNs. For example, \cite{bartlett2017} and \cite{neyshabur2017pac} developed the generalization bounds for FNNs based on the empirical Rademacher complexity under the Probably Approximately Correct (PAC)-Learning and PAC-Bayes frameworks, respectively. There are only a few studies that have investigated the Rademacher complexity bound of RNNs due to technical challenges. \cite{tu2019understanding} developed a generalization bound based on the matrix 1-norm and Fisher-Rao norm of the weight matrices. \cite{zhang2018stabilizing} and \cite{chen2019} adopted the PAC-Bayes method to establish bounds based on the spectral norm which can be smaller than the matrix 1-norm.

In this paper, we establish a new generalization bound for vanilla RNNs based on the empirical Rademacher complexity. We provide a unified framework to calculate the Rademacher complexity of vanilla RNNs that can be applied to a variety of commonly used loss functions satisfying certain conditions, while \citep{zhang2018stabilizing} and \citep{chen2019} use the hinge loss and the ramp loss, respectively. Moreover, both  \citep{zhang2018stabilizing} and \citep{chen2019} use the PAC-Bayes method which requires assuming a prior distribution (normal distribution) on the parameter weights and the random perturbation to establish the generalization bounds. We obtain a tighter bound by utilizing the Lipschitz continuity of RNNs under the PAC-learning framework without assuming any prior distributions and employing a mathematical induction strategy by making use of the recurrent structure of RNNs. 

In addition to deriving the generalization bound of RNNs, we also establish a new estimation error bound for the RNN-based estimators obtained through empirical risk minimization (ERM) \citep{vapnik2013nature} based on the observed data, which has not been investigated in the literature. 
ERM is commonly used for training models in both regression and classification tasks with no assumption on the distribution of the outputs. The estimator obtained by minimizing the empirical risk on the training set is expected to be close to the oracle networks \citep{vapnik1999nature}. This procedure and its regularized versions have been studied in learning theory \citep{koltchinskii2011empirical}. \cite{alquier2019} provide estimation bounds and sharp oracle inequalities for global regularized estimators in linear models. Fast convergence rates can be derived from the bounds given in \citep{alquier2019} for several specific problems, including matrix completion, logistic LASSO (Least Absolute Shrinkage and Selection Operator) regression, and kernel regression. However, due to the complex nonlinear nature of the RNN classes, the estimation error bound of the RNN estimators obtained based on ERM remains unclear.

In this paper, we derive new estimation error bounds and the sharp oracle inequality for the RNN estimators obtained via ERM.  Given that RNNs have a nonlinear structure and the RNN-based estimator can be a local optimizer, we establish a sharp oracle inequality for the local RNN-based estimators obtained via ERM when the loss function satisfies a Bernstein condition \citep{bartlett2006,alquier2019}. We also provide a new generalization bound of the local Rademacher complexity of RNNs involved in the estimation error bound.  We show that the Frobenius norm distance between the RNN estimators and the oracle networks can be bounded by the same order of the local Rademacher complexity. Moreover, the estimated excess risk can be controlled in the same order as the square of the local Rademacher complexity. These estimation bounds serve as a theoretical basis for training RNNs via ERM in practice.

\section{Contributions and related work}\label{S5}

The contributions of our work are summarized as follows.
\begin{itemize}
    \item 
    We provide a general formulation for the generalization error bound of vanilla RNNs, which can be applied to a variety of commonly used loss functions, including hinge loss, ramp loss, and cross-entropy loss as special cases \cite{Tian2022}.

    \item For the hinge loss and the ramp loss used in \cite{zhang2018stabilizing} and \cite{chen2019}, respectively, based on the same assumption on the Frobenius and spectral norms of the weight matrices (mixed norms), we provide a tighter bound of vanilla RNNs under the PAC-learning framework without assuming any prior distributions, while \cite{zhang2018stabilizing} and \cite{chen2019} employ the PAC-Bayes which requires assuming a prior normal distribution on the parameters and the random perturbation to obtain the bounds.  Moreover, when a boundedness condition imposed by  \cite{chen2019} is assumed for the hidden layer activation function, we obtain an even sharper generalization bound than the one given in \cite{chen2019}. 

    \item We provide a generalization bound completely based on the Frobenius norms of the weight matrices without any constraint on the spectral norms, which have not been studied in the literature. It meets the immediate needs for seeking theoretical guidance from the practical perspective, as many RNN-based applications \citep{xie2017all,fiorillo2021deepsleepnet} add an $L_2$ norm penalty on the weights to avoid the exploding gradient problem.
    \item We establish a sharp estimation error bound for the RNN-based estimator obtained through ERM. We also provide a new generalization bound for the local Rademacher complexity that is involved in the estimation error bound. The sharp bound for the RNN-estimator is established under a Bernstein condition on the loss function. The estimation error bound is new for RNNs, and it has not been studied in the aforementioned papers which mainly focus on the generalization bound.
\end{itemize}

The contributions of our work compared to the existing ones are summarized in Table \ref{TT1}, where \usym{1F5F8} and \usym{2715} denote ``unavailable" and ``available", respectively. The detailed comparison of our generalization bounds to other existing bounds based on the mixed norms of the weight matrices is given in Section \ref{sec:comparision}. 

\begin{table}[!ht]
    \centering
    \caption{The contribution of our work.
    }\label{TT1}
     \resizebox*{135mm}{!}{
    \begin{tabular}{cccccc}
    \hline
        & & \cite{tu2019understanding} & \cite{zhang2018stabilizing} & \cite{chen2019} & Ours  \\ \hline
       Loss function &  & Hinge loss & Hinge loss & Ramp loss & Lipschitz loss (Hinge; \\& && & &ramp; cross-entropy) \\ Mixed norms 
       & ReLU & \usym{1F5F8} & \usym{1F5F8} & \usym{2715} & \usym{1F5F8} \\ 
       & Tanh & \usym{2715} & \usym{2715} & \usym{1F5F8} & \usym{1F5F8} \\ 
        Frobenius norm&  & \usym{2715} & \usym{2715} & \usym{2715} & \usym{1F5F8} \\ 
       Estimation Bound & & \usym{2715} & \usym{2715} & \usym{2715} & \usym{1F5F8} \\ \hline
    \end{tabular}
    }
\end{table}

The rest of the paper is organized as follows. Section \ref{notation} introduces the mathematical notations used in the paper. In Section \ref{S1}, we present the main results for the new generalization bound based on the empirical Rademacher complexity for vanilla RNNs. Section \ref{S2} provides the estimation error bound and the sharp oracle inequality for the RNN estimators. We evaluate the obtained bounds using three real-world datasets in Section \ref{S3}, and make concluding remarks in Section \ref{S4}. Due to the limited space, all technical proofs of the theoretical results are relegated to the Appendix.

\subsection{Notations}\label{notation}
For two sequences of positive numbers $\beta_{1n}$ and $\beta_{2n}$, $\beta_{1n}\ll \beta_{2n}$ implies that $\beta_{1n}=o(\beta_{2n})$ and $\beta_{1n}=O(\beta_{2n})$ means that there exist a constant $C  \in (0,\infty)$ and a positive integer $n_{0} \geq 1$ such that  $\beta_{1n} \leq C\beta_{2n}$ for $n \geq n_{0}$. $[f_{t}]_{k}$ represents the $k$-th component of the vector $f_{t}$. For a matrix $A=\{a_{ij}, i=1,2,...,n;j=1,2,...,d.\}$, $\|A\|_{F}$ denotes the Frobenius norm; $\|A\|_{\infty}$ denotes the maximum absolute value of the entries in $A$; $\|A\|_{1}$ denotes the 1-norm with $\|A\|_{1} =\underset{j}{\max}\{\sum_{i}|a_{ij}|\}$. We denote the spectral norm by $\|A\|_{\sigma}$ as the largest singular value of the matrix $A$. For a vector $v$, $\|v\|_{F}$ denotes the Euclidean norm. 

\begin{table*}[!ht]
\caption{Table of mathematical notations}%
\centering\resizebox*{135mm}{!}{\
\begin{tabular}{cccc}
\hline
Notations & Descriptions & Notations & Descriptions\\ \hline
$n$ & sample size & $t$ & sequence length  \\
$d_{x}$ & embedding size & $d_{h}$ & dimension of hidden layer \\
$d_{y}$ & dimension of the output & $d$ & maximum among $\{d_{x}, d_{h}, d_{y}\}$ \\
$x_{t}$ & inputs at $t$-th time-step & $X_{t}$ & concatenating $x_{1},...,x_{t}$ \\
$h_{t}$ & output of hidden layer & $y_{t}$ & output at $t$-th time-step \\
$z_{t}$ & label of $X_{t}$ & $f_{t}$ & a mapping from inputs to output \\
$\sigma_{h}$ & activation operator of hidden state & $U$ & weight matrix of hidden state \\
$W$ & weight matrix of inputs & $V$ & weight matrix of output \\
$\rho_{h}$ & Lipschitz parameter of $\sigma_{h}$ & $q$ & activation function mapping $f_{t}$ to response \\
$\mathcal{F}_{t}$ & function class & $\mathcal{E}$ & excess risk \\
$\ell$ & loss function & $K$ & number of classes \\
$\mathfrak{R}$ & Rademacher complexity & $\mathcal{R}$ & risk function\\  
\hline
\end{tabular}
}
\label{bb3}
\end{table*}

\section{Generalization bounds of vanilla RNNs}\label{S1}

In this section, we first introduce vanilla RNNs and the risk function in subsection \ref{S1-1}. Under the assumptions given in subsection \ref{S1-2}, we establish the empirical Rademacher complexity bounds of RNNs for unbounded and bounded activation operators, respectively, in subsections \ref{S1-4} and \ref{S1-5}, and we present the generalization error bound in subsection \ref{S1-6}. Lastly,  we discuss the Lipschitz continuity of the loss function in subsection \ref{S1-7}.

\subsection{Vanilla RNNs and the risk function}\label{S1-1}

A vanilla RNN is a type of neural network designed for processing sequential data. The fundamental building block of a vanilla RNN is the recurrent unit, which maintains a hidden state updated at each time step based on the current input and the previous hidden state. Given the previous hidden state $h_{i,t-1}$ and the current input $x_{i,t}$, the vanilla RNNs compute the hidden state $h_{i,t}$ and the output $y_{i,t}$ as:
\begin{equation}\label{E6}
  h_{i,t}=\sigma_{h}(Uh_{i,t-1}+Wx_{i,t}),  \; \text{and} \;  y_{i,t}=Vh_{i,t},
\end{equation}
\begin{itemize}
  \item $x_{i,t} \in \mathds{R}^{d_{x}}$ ;
  \item $h_{i,t} \in \mathds{R}^{d_{h}}$ with $h_{i,0}=0$;
  \item $U \in \mathds{R}^{d_{h}\times d_{h}}$,  $V \in \mathds{R}^{d_{y}\times d_{h}}$ and $W \in \mathds{R}^{d_{h}\times d_{x}}$;
  \item $d=\max\{d_{x},d_{y},d_{h}\}$,
\end{itemize}
where $U$, $V$ and $W$ are the weight matrices of the network. Equation \ref{E6} shows that the output of vanilla RNNs depends on both the input and the hidden state that stores past information observed in the sequence. The activation operator $\sigma_{h}$ is entry-wise, i.e., $\sigma_{h}([v_{1},...,v_{d}]^{T})=[\sigma_{h}(v_{1}),...,\sigma_{h}(v_{d})]^{T}$, and is Lipschitz continuous with Lipschitz continuity parameter $\rho_{h}$. 

We consider multi-class classification tasks with the labels $z_{t} =\{z_{i,t}\}_{i=1}^{n} \in \mathcal{Z}=\{1,...,K\}$. 
Given the input sequence $\{x_{1},...,x_{t}\}$, we define $X_{t}$ by concatenating $x_{1},...,x_{t}$ as columns of $X_{t}$. We denote $\mathcal{F}_{t}=\{f_{t}:X_{t}\rightarrow y_{t}\}$ as the class of mappings from the first $t$ inputs to the $t$-th output computed by vanilla RNNs. 

We next introduce the risk function which is the expected value of a loss function, and it is a key tool for evaluating a learning model. The risk function is  defined as
\begin{equation}\label{risk}
\mathcal{R}(f_{t})=\mathds{E}[\ell(q(f_{t}(X_{t})),z_{t})],
\end{equation}
where $\ell$ is the loss function, $q$ is the activation function mapping the output $f_{t}(X_{t})$ of vanilla RNNs to the response of interest, and $z_{t}$ is the label of $X_{t}$ at the time $t$. For notational simplicity, denote $\ell_{f_{t}} := \ell(q(f_{t}(X_{t})),z_{t})$ and $q_{t} := q(f_{t}(X_{t}))$. In practice, the activation function $q$ can have different functional forms for different learning tasks. For example, when the cross-entropy loss function is applied for a multi-class classification task, the softmax function is used such that
$q_{t}=\{q_{t,k}\}_{k=1}^{K}=\left\{\frac{e^{f_{t,k}}}{\sum_{k=1}^{K}e^{f_{t,k}}}\right\}_{k=1}^{K}$ with $f_{t,k}=[f_{t}(X_{t})]_{k}$. 

Let $\{X_{it},z_{it}\}_{i=1}^{n}$ be a random sample with size $n$ generated from the same distribution of $\{X_{t},z_{t}\}$.
The empirical risk function based on the random sample is defined as
 \begin{equation}\label{empiricalrisk}
    \widehat{\mathcal{R}}(f_{t})=\frac{1}{n}\sum_{i=1}^{n}\ell\left(q(f_{t}(X_{i,t})),z_{i,t}\right).
 \end{equation}
 We denote the class for the loss function by $\mathcal{F}_{\ell,t}=\{(X_{t},z_{t}) \rightarrow \ell_{f_{t}}: f_{t} \in \mathcal{F}_{t}\}$.  
 Given the sequence length $t$,
for a function $f_{t}\in \mathcal{F}_{t}$, denote its maximum by $\|f_{t}\|_{\infty} = \sup_{X_{t}} |f_{t}(X_{t})|$; for a loss function $\ell_{f_{t}}\in \mathcal{F}_{\ell,t}$, denote its maximum by
$\|\ell_{f_{t}}\|_{\infty} = {\sup}_{(X_{t},z_{t})}|\ell\left(q_{t},z_{t}\right)|$.

\subsection{Some assumptions}\label{S1-2}
To establish the generalization bound of the vanilla RNNs, we impose the following mild assumptions.
\begin{assumption}\label{A1}
    The Frobenius norm of the input data is bounded, i.e., $\|x_{t} \|_{F} \leq B_{x}$, for all  $t=1,2...$.
\end{assumption}

\begin{assumption}\label{A2}
    The Frobenius norms of weight matrices are bounded respectively, i.e., $\|U\|_{F} \leq B_{U}$, $\|V\|_{F} \leq B_{V}$ and $\|W\|_{F} \leq B_{W}$.
\end{assumption}

\begin{assumption}\label{A3}
   The activation operator $\sigma_{h}$ is Lipschitz continuous with parameter $\rho_{h}$ and $\sigma_{h}(0)=0$. 
\end{assumption}

\begin{assumption}\label{A5}
\begin{itemize}
    \item [(i)] 
    There exists a constant $C_{t}>0$ such that for any $f_{t} \in \mathcal{F}_{t}$, $\|\ell_{f_{t}}\|_{\infty} \leq C_{t}$;
    \item [(ii)] The loss function $\ell_{f_{t}}$ is $\rho$-Lipschitz with respect to $f_{t}$. 
\end{itemize}
\end{assumption}

\textbf{Remark:} Assumptions \ref{A1}-\ref{A3} are moderate assumptions. Specifically,  Assumption \ref{A1} is a fundamental assumption about the input data, which is also assumed in the existing theoretical works on RNNs \citep{tu2019understanding,zhang2018stabilizing,chen2019}. In Assumption \ref{A2}, we use the Frobenius norms to limit the upper bounds of the weight matrix parameters in which the capacity of the networks is proportional to each element. Therefore, when training on true labels, we expect that the model can capture the functional dependence between the input and output data, so we can see how each element in the weight matrix affects the upper bound of the generalization error. Compared to the bounds assumed on the spectral norm of the weight matrices \citep{zhang2018stabilizing,chen2019}, the bounds imposed only on the Frobenius norm can liberate the RNN model class.

Assumption \ref{A3} holds for most commonly used activation operators, such as $\tanh(\cdot)$ and $\text{ReLU}(\cdot)$. Compared to Assumption 3 given in \cite{chen2019}, our assumption does not need to impose an upper bound of the activation operator $\sigma_{h}$. As a result, our theory can be applied to a wider range of activation operators including $\text{ReLU}(\cdot)$. Assumption \ref{A5} (i) imposes an upper bound for the loss function. This assumption is needed in order to apply McDiarmid’s inequality to bound the stochastic error in the generalization error for a machine learning method \cite{mohri2018}. Assumption \ref{A5} (ii) is required to calculate the Rademacher complexity of the functional class of the loss functions. In practice, the most commonly used loss functions satisfy this assumption. For example, both the hinge loss and the cross-entropy loss are $\sqrt{2}$-Lipschitz with respect to $f_{t}$. We provide a tight generalization error bound for vanilla RNNs when the loss function satisfies the two conditions given in Assumption \ref{A5} (i) and (ii). 

\subsection{Related concepts and lemmas}\label{S1-3}

Rademacher complexity is a commonly used tool to derive data-dependent upper bounds on the learnability of function classes and a good measure of the richness of a class of real-valued functions with respect to a probability distribution. Below we give the definition of empirical Rademacher complexity.

\begin{definition}
    Let $\mathcal{H}$ be a function class mapping from $\mathcal{S}$
to $[a,b]$ and $S=\{s_{1},...,s_{n}\}$ be a fixed sample of size $n$ with elements in $\mathcal{S}$. The empirical Rademacher complexity of $\mathcal{H}$ given the sample $S$ is defined as
    \begin{equation*}
    \mathfrak{R}_{S}(\mathcal{H})=\mathds{E}_{\epsilon}\left[\underset{h \in \mathcal{H}}{\sup}\;\frac{1}{n}\sum_{i=1}^{n} \epsilon_{i}h(s_{i})\right],
    \end{equation*}
    where $(\epsilon_{i})_{i=1}^{n}$ are i.i.d. Rademacher random variables taking values in $\{-1,1\}$. (i.e., $\mathds{P}(\epsilon_{i}=1)=\mathds{P}(\epsilon_{i}=-1)=1/2$).
\end{definition}

\textbf{Remark:} 
The empirical Rademacher complexity measures on average how well the function class $\mathcal{H}$ correlates with random noise on $\mathcal{S}$. Unlike the VC dimension, Rademacher complexity is not restricted to binary functions. Moreover, it can provide a more refined measure of model complexity than the VC dimension, as it takes the data information into account. 

The generalization error bound of RNNs can be well-controlled by the empirical Rademacher complexity and the stochastic error term. We provide an empirical Rademacher complexity (ERC)-based generalization bound, which is presented in the following lemma.

\textcolor{black}{
\begin{lemma} \label{L88}
Under Assumption \ref{A5} (i), given a test sequence $(X_{t},z_{t})$ and the predicted label $\hat{z}_{t}$ for the input $X_{t}$, for any $\delta>0$, with probability at least  $1-\delta$ over the sample $S=\{X_{it},z_{it}\}_{i=1}^{n}$, 
for any $f_{t}\in \mathcal{F}_{t}$, one has
    \begin{equation}\label{Loss1}
        \mathds{P}[ \hat{z}_{t} \neq z_{t}] \leq \mathcal{R}(f_{t}) \leq \widehat{\mathcal{R}}(f_{t})+2\mathfrak{R}_{S}(\mathcal{F}_{\ell,t})+3C_{t}\sqrt{\frac{\log(2/\delta)}{2n}}.
    \end{equation}
\end{lemma}
}

\textbf{Remark:} 
We extend the results of \cite{mohri2018} and \cite{scott2014rademacher} to obtain Lemma \ref{L88} and provide the proof of  Lemma \ref{L88} in Appendix \ref{S8-1-5}. Lemma \ref{L88} shows that the difference between the population and the empirical risk functions can be bounded by two terms -- the Rademacher complexity and the stochastic error bound. We provide a tight bound for the Rademacher complexity of the vanilla RNNs based on PC-learning where covering numbers will be used to directly upper bound this complexity term. The bound for the stochastic error term can be obtained by applying McDiarmid's inequality under Assumption \ref{A5} (i).

Our theoretical analysis within the framework of PAC-learning consists of two steps. First, we characterize the Lipschitz continuity of the vanilla RNNs with respect to the model parameters. Second, we bound the covering number of the function class $\mathcal{F}_{\ell,t}$. To facilitate presentation, we denote
\begin{eqnarray}
a =\rho _{h}B_{x},b_{t}=\frac{t(\rho _{h}B_{U})^{t-1}-c_{t}}{\rho
_{h}B_{U}-1},c_{t}=\frac{(\rho _{h}B_{U})^{t}-1}{\rho _{h}B_{U}-1}, \\
g_{t} =\rho _{h}B_{x}B_{V}B_{W}\max\{c_{t},\rho _{h}B_{U}b_{t}\}.
\label{EQ:abcgt}
\end{eqnarray}%
If $\Vert U\Vert _{\sigma }\leq M_{U}$, we denote
\begin{eqnarray}
b_{t}^{\prime }=\frac{t(\rho _{h}M_{U})^{t-1}-c_{t}^{\prime }}{\rho
_{h}M_{U}-1},c_{t}^{\prime }=\frac{(\rho _{h}M_{U})^{t}-1}{\rho _{h}M_{U}-1},\\
g_{t}^{\prime }=\rho _{h}B_{x}B_{V}B_{W}\max\{c_{t}^{\prime },\rho
_{h}B_{U}b_{t}^{\prime }\}.
\label{EQ:bcgtprime}
\end{eqnarray}

\subsection{The Rademacher complexity for unbounded activation operators}\label{S1-4}

First of all, we calculate the empirical Rademacher complexity in terms of $\mathfrak{R}_{S}(\mathcal{F}_{\ell,t})$ where the hidden layer activation operator $\sigma_{h}$ is not bounded, such as ReLU and Leaky ReLU.

\begin{theorem}\label{T1} 
\textbf{The Rademacher complexity of vanilla RNNs:}

(a) Under Assumptions \ref{A1}-\ref{A3} and \ref{A5} (ii), given the sample $S=\{(X_{it},z_{it})_{i=1}^{n}\}$, the empirical Rademacher complexity $\mathfrak{R}_{S}(\mathcal{F}_{\ell,t})$ satisfies 
    \begin{equation*}
        \mathfrak{R}_{S}(\mathcal{F}_{\ell,t}) = O\left(\frac{\rho dg_{t}\sqrt{\log(d)}\log\left(\rho_{h}nB_{x}B_{V}B_{W}c_{t}\right)}{n}\right).
    \end{equation*}
    
(b) If $\|U\|_{\sigma} \leq M_{U}$, then we obtain a tighter bound for the empirical Rademacher complexity $\mathfrak{R}_{S}(\mathcal{F}_{\ell,t})$ given as
    \begin{equation}\label{E16}
       \mathfrak{R}_{S}(\mathcal{F}_{\ell,t}) = O\left(\frac{\rho dg_{t}^{\prime}\sqrt{\log(d)}\log\left(\rho_{h}nB_{x}B_{V}B_{W}c_{t}^{\prime}\right)}{n}\right).
    \end{equation}
\end{theorem}

\textbf{Remark for Theorem \ref{T1}:} \textcolor{black}{
We provide a unified framework to calculate the Rademacher complexity $\mathfrak{R}_{S}(\mathcal{F}_{\ell,t})$ of vanilla RNNs for any $\rho$-Lipschitz loss functions, including the hinge loss, ramp loss and cross-entropy loss as special cases. We show in Appendix \ref{S8-3} that the hinge and cross-entropy loss functions are $\sqrt{2}$-Lipschitz ($\rho=\sqrt{2}$), and it has been proven in Lemma A.3 of \cite{bartlett2017} that the ramp loss is $2/\gamma$-Lipschitz ($\rho=2/\gamma$) with respect to the argument $f_{t}$, where $\gamma$ is the margin value of the ramp loss. 
Our proof exploits the Lipschitz continuity of the loss function $\ell_{f_{t}}$, uses a mathematical induction strategy by taking advantage of the recurrent structure of the vanilla RNNs, and applies the covering number theory from Lemma 3.2 in \cite{bartlett2017} and Dudley's entropy integral to establish Theorem \ref{T1}. The detailed proof of Theorem \ref{T1} is provided in Appendix \ref{S8-1-6}.
}

In Theorem \ref{T1} (a), we first derive the Rademacher complexity completely based on the Frobenius norms of the weight matrices, while the existing bounds in the literature such as the one given in \cite{chen2019} are based on the mixture of both spectral and Frobenius norms. One advantage of the result given in (a) is that it does not require any constraints on the spectral norms, so it makes the optimization procedure for parameter estimation much easier \citep{huang2018new,liu2021robust}. In practice, it is more difficult to carry out an optimization procedure with a constraint or penalty on the spectral norm than that on the Frobenius norm.
Moreover, the bound in (a), which is built on the Frobenius norms, better meets the needs for seeking theoretical guidance from the practical perspective, as many RNN-based applications \citep{xie2017all,fiorillo2021deepsleepnet} add an $L_2$ norm penalty term on the weight matrices to the loss function to avoid the exploding gradient problem \citep{goodfellow2016deep}. In Theorem \ref{T1} (b), with an additional assumption on the spectral norm of $U$, we establish a tighter bound for the empirical Rademacher complexity. This bound is tighter than the one in Theorem \ref{T1} (a).

To be specific, if the spectral norm of $U$ is bounded with $\|U\|_{\sigma} \leq M_{U}$, the $b_{t}$ and $c_{t}$ under Assumptions \ref{A1}-\ref{A3} can be directly updated to the spectral norm of $U$ with $b_{t}^{\prime}=\frac{t(\rho_{h}M_{U})^{t-1}-c_{t}^{\prime}}{\rho_{h}M_{U}-1}$ and $c_{t}^{\prime}=\frac{(\rho_{h}M_{U})^{t}-1}{\rho_{h}M_{U}-1}$. Therefore, the Rademacher complexity can be tighter than the result based on the Frobenius norm of $U$. Of course, the cost of obtaining a tighter bound is that we have assumed a stronger condition on the weight matrix $U$, which further limits the size of the function class. The argument why we can directly update the Frobenius norm $B_{U}$ to $M_{U}$ is in light of $\|A_{1}A_{2}\|_{F} \leq \|A_{1}\|_{\sigma}\|A_{2}\|_{F}$ in which the spectral norm of $\|A_{1}\|_{\sigma}$ is bounded. For detailed proof, please see Lemma \ref{L2} (b) and Lemma \ref{L3} (b) in Appendix \ref{S8-1-6}.

In addition, our bound does not require that the activation operator $\sigma_{h}$ at the hidden layer is bounded above by a constant, which is assumed in \cite{chen2019}. This relaxation makes our bound applicable to a number of commonly used activation operators, including $\text{ReLU}(\cdot)$, $\text{PReLU}(\cdot)$ and Leaky $\text{ReLU}(\cdot)$. 

\subsection{The Rademacher complexity for bounded activation operators}\label{S1-5}

Next, we derive the bound for the empirical Rademacher complexity of vanilla RNNs in the case where the activation function of the hidden layer is bounded, satisfying the condition given below.

\setcounter{assumption}{0} 
\renewcommand{\theassumption}{A}

\begin{assumption}\label{A7}
    The hidden layer activation operator $\sigma_{h}$ is entry-wise bounded by $b$. 
\end{assumption}

In Assumption \ref{A7}, we assume that the activation operator $\sigma_{h}$ is bounded above by a constant. This assumption holds for certain activation operators, such as $\tanh$ and sigmoid functions, but not for $\text{ReLU}$ and Leaky $\text{ReLU}$. Under Assumption \ref{A7} together with the conditions given in  Theorem \ref{T1} (b), we can obtain a tighter bound given below for the empirical Rademacher complexity of vanilla RNNs than the one given in Theorem \ref{T1} (b).

\textcolor{black}{
\begin{corollary}\label{L9}
Under the same conditions in Theorem \ref{T1} (b) together with Assumption \ref{A7}, we can obtain a tighter bound for vanilla RNNs compared to Equation (\ref{E16}) in Theorem \ref{T1}, given as
    \begin{equation}\label{E20}
       \mathfrak{R}_{S}(\mathcal{F}_{\ell,t}) = O\left(\frac{\rho dg_{t}^{*}\sqrt{\log(d)}\log\left(nB_{V}\min\{b\sqrt{nd},\rho_{h}B_{x}B_{W}c_{t}^{\prime}\}\right)}{n}\right),
    \end{equation}
where $g_{t}^{*}=B_{V}\min\{b\sqrt{nd},\rho _{h}B_{x}B_{W}c_{t}^{\prime}\}\max\{1,\rho
_{h}B_{U}b_{t}^{\prime}/c_{t}^{\prime}\}$.
\end{corollary}
}
\textbf{Remark for Corollary \ref{L9}:} If we assume that the activation function $\sigma_{h}$ is bounded, the Rademacher complexity given in Equation (\ref{E20}) is tighter than the one given in Equation (\ref{E16}) in Theorem \ref{T1}. Moreover, the bound in Equation (\ref{E20}) shows that for $\rho_{h}M_U>1$,
when the sequence length $t$ is large satisfying $c_{t}^{\prime}\geq b\sqrt{nd}/(\rho_{h}B_{x}B_{W})$,  we should consider using a bounded activation operator such as tanh, as the leading term $g_{t}^{*}$ of the bound given in (\ref{E20}) becomes $g_{t}^{*}=B_{V}b\sqrt{nd}\max\{1,\rho_{h}B_{U}b_{t}^{\prime}/c_{t}^{\prime}\}$ which does not increase rapidly with $t$. However, the leading term $g_{t}^{\prime}$ of the bound in Equation (\ref{E16}) for unbounded activator operators such as ReLU can grow exponentially with the $t$ value.

In the literature, \cite{zhang2018stabilizing} and \cite{chen2019} provide the bounds for the empirical Rademacher complexity when the ramp loss function is used. \cite{chen2019} have shown that their bound can be tighter than the one given in \cite{zhang2018stabilizing} when the activation function is bounded by a constant. We demonstrate a comparison of our bound (Equation (\ref{E20}) with $\rho=2/\gamma$ for the ramp loss) with the one (Equation (3) of Theorem 3) given by \cite{chen2019} and other existing bounds in Section \ref{S3}. Our bound provided in Equation (\ref{E20}) of Corollary \ref{L9} involves one less $M_{U}^{t}$ than the one given in  Equation (3) of \cite{chen2019}. As the spectral norm \textcolor{black}{$M_{U}^{t}$} is typically greater than one, their bound is larger than ours especially when $t$ grows. 

Moreover, both \cite{zhang2018stabilizing} and \cite{chen2019} adapt the PAC-Bayes framework in \citep{neyshabur2017pac}  that requires a prior normal distribution on the parameters to obtain the bound, while our proof which does not require assuming a prior distribution relies on the Maurey sparsification lemma \citep{pisier1981remarques}, and its use here is inspired by covering number bounds for linear function classes \citep{zhang2002covering}. We obtain a new bound of the order $O\left(\frac{\log(n)}{n}\right)$ in terms of the sample size $n$, compared to the bound $O\left(\frac{1}{\sqrt{n}}\right)$ given in \cite{chen2019}.

\subsection{The generalization bounds of vanilla RNNs}\label{S1-6}

By Lemma \ref{L88} and Theorem \ref{T1}, we can immediately obtain the generalization error bounds of vanilla RNNs given below.

\begin{theorem}\label{T2}
Let the activation operator $\sigma_{h}$ be given, and Assumptions \ref{A1}-\ref{A5} hold. Then, we consider multi-class classification tasks $\{X_{t},z_{t}\}$ with the label $z_{t} \in \mathcal{Z}=\{1,...,K\}$ and $S=\{(X_{it},z_{it})_{i=1}^{n}\}$, with probability at least $1-\delta$ over $S$, for sufficiently large sample size $n$, for any $f_{t}\in \mathcal{F}_{t}$,

(a) we have
    \begin{equation*}
        \begin{split}
        \mathds{P}[\hat{z}_{t} \neq z_{t}] \leq \widehat{\mathcal{R}}(f_{t})
         +O\left(\frac{\rho dg_{t}\sqrt{\log(d)}\log(\rho_{h}nB_{x}B_{V}B_{W}c_{t})}{n}\right)
         +3C_{t}\sqrt{\frac{\log(2/\delta)}{2n}},
        \end{split}
    \end{equation*}
    
(b) if $\|U\|_{\sigma} \leq M_{U}$, we have
    \begin{equation}\label{E17}
        \begin{split}
            \mathds{P}[\hat{z}_{t} \neq  z_{t}] \leq \widehat{\mathcal{R}}(f_{t})+O\left(\frac{\rho dg_{t}^{\prime}\sqrt{\log(d)}\log\left(\rho_{h}nB_{x}B_{V}B_{W}c_{t}^{\prime}\right)}{n}\right)
            +3C_{t}\sqrt{\frac{\log(2/\delta)}{2n}}.
        \end{split}
    \end{equation}
\end{theorem}  

In Theorem \ref{T2}, we provide a general formulation for the generalization error bound of vanilla RNNs when the loss function is $\rho$-Lipschitz and is bounded by $C_{t}$. The generalization bound depends on the total number of weights, the sample size $n$, and the range of $\rho_{h}B_{U}$.
From Theorem \ref{T2}, we can see that with proper normalization on the model parameters, the model complexity of vanilla RNNs does not suffer significantly from the dimensionality problem.  We also highlight a tradeoff between the generalization ability and the representation ability of vanilla RNNs. We can observe that when $\rho_{h}B_{U}$ is strictly smaller than 1, the generalization bound is nearly independent of $t$. However, the hidden state only has limited representation ability due to its magnitude diminishing as $t$ increases. On the contrary, when $\rho_{h}B_{U}$ is strictly greater than 1, the representation ability can be enhanced but the generalization ability becomes worse.

\subsection{Discussions on the loss function}\label{S1-7}

A loss function often has a natural bound because of truncation. For example, the ramp loss is a truncated version of the hinge loss, so it is bounded by $C_t=1$. For some loss functions such as the hinge loss and the cross-entropy loss, they do not have a natural bound. Then, we can impose the following condition.  

\begin{condition} (Boundedness condition) \label{A6}
    There exists a constant $\omega_{t} >0$ such that for any $f_{t} \in \mathcal{F}_{t}$, $\|f_{t}\|_{\infty} \leq \omega_{t}$.
\end{condition}

\textbf{Remark: }When a loss function does not have a natural bound, imposing an upper bound for the output function $f_t$ given in Condition \ref{A6} is a typical operation in the existing works
\cite{alquier2019,cao2020generalization}. In practice, one often has to either truncate the loss function $\ell_{f_{t}}$ itself such as the ramp loss or put a constraint on the output function $f_t$ to control the stochastic error given in  Lemma \ref{L88}.

\begin{lemma}\label{Loss2}
Under Assumption \ref{A5} (ii) and Condition \ref{A6} with $\ell_{f_{t}}(0,z_{t})=0$, one has
\begin{equation}
   \|\ell_{f_{t}}\|_{\infty} \leq C_{t} = 2\rho \omega_{t}.
\end{equation}
\end{lemma}

The proof of Lemma \ref{Loss2} is provided in subsection \ref{S8-1-0}. Lemma \ref{Loss2} shows that for a loss function that does not possess a natural upper bound, when the output function $f_t$ from the RNNs is bounded by $\omega_t$ together with the $\rho$-Lipschitz continuity property, the loss function can be bounded above by $2\rho \omega_{t}$. As a result, the result in Theorem \ref{T2} holds for this setting with $C_{t} = 2\rho \omega_{t}$. 

Theorem \ref{T2} establishes a general framework for the generalization error bound that can be applied to a variety of commonly used loss functions, including the cross-entropy loss, the hinge loss, and the ramp loss.

Specifically, the cross-entropy loss is defined as 
\begin{equation*}
    \ell_{f_{t}}=\ell(q(f_t(X_t)),z_{t})=-\sum\nolimits_{k=1}^{K}I(z_{t}=k)\log(q_{t,k}),
\end{equation*}
where $q_{t}=\{q_{t,k}\}_{k=1}^{K}$ is the soft-max function with $q_{t,k}=\frac{e^{f_{t,k}}}{\sum_{k=1}^{K}e^{f_{t,k}}}$ and $f_{t,k}=[f_{t}(X_{t})]_{k}$. The cross-entropy does not have a natural bound, but it is $\rho$-Lipschitz with $\rho=\sqrt{2}$. The proof for the Lipschitz continuity of the cross-entropy loss is provided in Appendix \ref{S8-3}.
Together with Condition \ref{A6}, the upper bound for the cross-entropy loss is $C_{t}=2\sqrt{2}\omega_{t}$. 

Denote the margin operator $$\psi_{f_{t}} :=\psi\left(q(f_{t}(X_{t})),z_{t}\right)=\underset{k \neq z_{t}}{\max}\left[q(f_{t}(X_{t}))\right]_{k}-\left[q(f_{t}(X_{t}))\right]_{z_{t}},$$ where $q$ is an activation operator and it is Lipschitz with parameter $\rho_q$ with $q(0)=0$. Without loss of generality, let $\rho_q=1$. Then, the hinge loss function is defined as 
$\ell_{f_{t}}=\max\{0,1+\psi_{f_{t}}\}$. Similar to the cross-entropy loss, the hinge loss is $\rho$-Lipschitz with $\rho=\sqrt{2}$ for multiclass classification; see the proof in Appendix \ref{S8-3}. Together with Condition \ref{A6}, the upper bound for the hinge loss is also $C_{t}=2\sqrt{2}\omega_{t}$, although it does not possess a natural bound either. 

The ramp loss used in \citep{zhang2018stabilizing,chen2019} is defined as 
\begin{equation*}
    \ell_{f_{t}}=I(\psi_{f_{t}}>0)+(1+\psi_{f_{t}}/\gamma)I(-\gamma \leq \psi_{f_{t}} \leq 0),
\end{equation*}
where $\gamma$ is its margin value.  The ramp loss can be viewed as a truncated version of the aforementioned hinge loss, and it has a natural bound with $C_t=1$ and it is $\rho$-Lipschitz with $\rho=2/\gamma$. The proof for the Lipschitz continuity of the ramp loss is given in Lemma A.3 of \cite{bartlett2017}.

In this section, we have derived the generalization error bounds for vanilla RNNs based on the empirical Rademacher complexity when the loss function is $\rho$-Lipschitz. In the next section, we will present a new estimation error bound for the RNN estimators obtained via ERM.

\section{Estimation error bounds of vanilla RNNs}\label{S2}

In this section, we investigate the estimation error bound of the vanilla RNNs-based estimator obtained through empirical risk minimization (ERM) when the output function $f_{t}$ is bounded and the loss function satisfies a Bernstein condition. The estimation error measures the error of an estimator with respect to the infimum of the errors achieved by a fixed class of functions \citep{mohri2018}. Establishing an upper bound for the estimation error of an estimator obtained from a machine learning algorithm is important to evaluate the performance of this learning algorithm.

We denote $\mathcal{\widetilde{F}}_{t}=\{f_{t}:X_{t}\rightarrow y_{t}\}$ as the class of mappings from the first $t$ inputs to the $t$-th output computed by vanilla RNNs  satisfying Assumptions \ref{A1}-\ref{A3}. The ultimate goal of a learning algorithm is to find a hypothesis (the best-in-class hypothesis) among a fixed class of functions such that the risk function is minimized \citep{vapnik1991principles}. In our considered RNNs setting, this best-in-class hypothesis, denoted by $f_{t}^{*}$, is defined as:  
\begin{equation}\label{fstar}  
f_{t}^{*} \in \underset{f_{t}\in \mathcal{\widetilde{F}}_{t}}{\text{argmin}}\; \mathcal{R}(f_{t}),
\end{equation}
where the risk function $\mathcal{R}(f_{t})$ is defined in (\ref{risk}), so we see that  $f_{t}^{*}$ is the best-in-class hypothesis among all functions in the RNN class $\mathcal{\widetilde{F}}_{t}$ at which the risk function $\mathcal{R}(f_{t})$ is minimized.

In general, the risk $\mathcal{R}(f_{t})$ cannot be computed because the population distribution of the data is unknown to the learning algorithm. Instead, we can compute the empirical risk $\widehat{\mathcal{R}}(f_{t})$ by computing the average of the loss function over the training set given in (\ref{empiricalrisk}). Then we can obtain an estimator of $f_{t}^{*}$ by minimizing $\widehat{\mathcal{R}}(f_{t})$, i.e., this estimator is obtained through ERM. In practice, it is difficult to obtain the global minimizer of $\widehat{\mathcal{R}}(f_{t})$ over all functions belonging to the RNN class $\widetilde{F}_{t}$. Instead, we can obtain an RNN estimator $\widehat{f}_{t}$ that is a local minimizer of $\widehat{\mathcal{R}}(f_{t})$  and  satisfies:
\begin{equation}\label{RNNest}  
\mathcal{\widehat{R}}(\widehat{f_{t}}) \leq \underset{f_{t}\in \mathcal{\widetilde{F}}_{t}}{\min}\; \mathcal{\widehat{R}}(f_{t})+\Delta_{t}.
\end{equation}
The sequence $\Delta_{t}$ measures the difference between the empirical risk of  $\widehat{f}_{t}$ and the global minimum over all networks in the RNN class. Notice that $\Delta_{t} \geq 0$ and $\Delta_{t}=0$ if $\widehat{f}_{t}$ is the global empirical risk minimizer. Note that we do not require the RNN estimator $\widehat{f}_{t}$ defined above to be the global minimizer of the empirical risk, which is difficult to obtain in practice. However, the difference in their empirical risk values should be bounded by $\Delta_{t}$.

To derive the estimation error bound for the RNN-based estimator given in (\ref{RNNest}), we first need to obtain an upper bound for the local Rademacher complexity \citep{bartlett2005local}, which is involved in the estimation error bound. Different from the global Rademacher complexity that is computed over the entire RNN class, the local Rademacher complexity is calculated over the intersection of the RNN class $\mathcal{F}_{t}$ and a ball of radius r and with the center $f_t^{*}$. The theorem below provides an upper bound for the local Rademacher complexity. 

\begin{theorem}\label{T3}
Under Assumptions \ref{A1}-\ref{A3}, for any $c_{1}>1$ and $\alpha >0$, let $rB_{F}$ denote the radius $r$ ball based on the Frobenius norm, that is, $rB_{F}=\{f_{t} \in \mathcal{F}_{t}: \|f_{t}-f_{t}^{*}\|_{F} \leq r\}$,

(a) the local Rademacher complexity with $rB_{F}$ has an upper bound: 
    \begin{equation}\label{E5}
     \mathfrak{R}_{S}(rB_{F}) \leq \frac{\eta_{t}}{\sqrt{n}}r,
    \end{equation}
    
(b) if $\|U\|_{\sigma} \leq M_{U}$, we have
    \begin{equation*}
        \mathfrak{R}_{S}(rB_{F}) \leq \frac{\eta_{t}^{\prime}}{\sqrt{n}}r,
    \end{equation*}
where $\eta _{t} =\frac{3\sqrt{3}c_{1}}{\alpha }d\sqrt{\log (2d^{2})%
}g_{t}$,\;$\eta _{t}^{\prime } =\frac{3\sqrt{3}c_{1}}{\alpha }d\sqrt{%
\log (2d^{2})}g_{t}^{\prime }$, $g_{t}$ and $g_{t}^{\prime }$ are
defined in (\ref{EQ:abcgt}) and (\ref{EQ:bcgtprime}). 
\end{theorem}
The intuition for the proof of Theorem \ref{T3} is as follows. We perform an entropy integral on the covering numbers of the function class $\mathcal{F}_{t}$ within $rB_{F}$ and apply certain oracle inequalities to obtain the bound for the local Rademacher complexity. The detailed proof of Theorem \ref{T3} is presented in Appendix \ref{S8-2-2}. Compared to the global Rademacher complexity, it imposes some restricted conditions on the networks, so that its complexity is located within a specified area, rather than in the entire space.

We next introduce the excess risk $\mathcal{E}(f_{t})$ for $f_{t} \in \mathcal{\widetilde{F}}_{t}$, defined as
$\mathcal{E}(f_{t})=\mathds{E}(\mathcal{L}_{f_{t}})=\mathcal{R}(f_{t})-\mathcal{R}(f_{t}^{*})$, where $\mathcal{L}_{f_{t}}=\ell(q(f_{t}(X_{t})),z_{t})-\ell(q(f_{t}^{*}(X_{t})),z_{t})$. The excess risk measures the difference between the risk of a function and the optimal function given in (\ref{fstar}), and is also used to evaluate the performance of the RNNs estimator. We assume that the excess risk satisfies the Bernstein condition given below.
\textcolor{black}{
\begin{condition} (Bernstein condition).\label{A4}
There exists a constant $A_{t}>0$ such that for any
$f_{t} \in \mathcal{\widetilde{F}}_{t}$,
\begin{equation}\label{E12}
\mathds{E}[\|f_{t}-f_{t}^{*}\|_{F}^{2}] \leq A_{t} \mathcal{E}(f_{t}).
\end{equation}
\end{condition}
}

\textbf{Remark:} The Bernstein condition from \cite{bartlett2006} is a key assumption and has been widely used in statistical learning theory to guarantee a sharp estimation error bound for the estimator obtained via ERM \citep{alquier2019,Geoffrey2020}. It quantifies the identifiability or curvature of the excess risk function around its minimizer $f_{t}^{*}$ to ensure that the ERM estimator has a fast convergence rate.
 
The Bernstein condition given in Condition \ref{A4} is satisfied by commonly used loss functions, including the logistic loss and the hinge loss as discussed in \cite{alquier2019}. 
Under Condition \ref{A4}, we establish a sharp estimation error bound for the RNN-based estimator obtained from (\ref{RNNest}) given below.


\begin{theorem}\label{T4}
    (a) Assume that Assumptions \ref{A1}-\ref{A3}, Assumption \ref{A5} (ii) and Conditions \ref{A6}-\ref{A4} hold. For some positive constant $\theta_{t}$ such that $\rho A_{t}\theta_{t} <1$. Let $\vartheta_{t}=n\nu_{t}\theta_{t}^{2}(\varphi_{t}^{*})^{2}$ with \textcolor{black}{$\nu_{t}=\min\left\{\frac{1}{288},\frac{1}{207\theta_{t}\omega_{t}}\right\}$}. For $\Delta_{t}=o\left((\varphi_{t}^{*})^{2}\right)$ and $\omega_{t}=o(\sqrt{\eta_{t}})$, with probability larger than
    \begin{equation}\label{E2} 
    1-\frac{2\exp(-\vartheta_{t})}{1-\exp(-3\vartheta_{t})},
    \end{equation}
    we have
    \begin{equation*}
    \|\widehat{f_{t}}-f_{t}^{*}\|_{F} \leq \varphi_{t}^{*}\; \text{and}\;\mathcal{E}(\widehat{f_{t}}) \leq \rho \theta_{t}(\varphi_{t}^{*})^{2},
    \end{equation*}
    where
    \begin{equation}\label{E1} 
    \varphi_{t}^{*}=\frac{48\eta_{t}}{\sqrt{n}\theta_{t}}.
    \end{equation}
    (b) If $\|U\|_{\sigma} \leq M_{U}$, we have
    \begin{equation*}
    \|\widehat{f_{t}}-f_{t}^{*}\|_{F} \leq \phi_{t}^{*}\; \text{and}\;\mathcal{E}(\widehat{f_{t}}) \leq \rho \theta_{t}(\phi_{t}^{*})^{2},
    \end{equation*}
    where
    \begin{equation*}
    \phi_{t}^{*}=\frac{48\eta_{t}^{\prime}}{\sqrt{n}\theta_{t}}.
    \end{equation*}
\end{theorem}

The intuition for the proof of Theorem \ref{T4} is as follows. We first find an event $\Omega_{t}$ and show that the Frobenius norm distance between the estimator $\widehat{f}_{t}$ and the target function $\|\widehat{f_{t}}-f_{t}^{*}\|_{F}$ is bounded above by $\phi_{t}^{*}$ on the event $\Omega_{t}$. We further show that the expected excess risk $\mathcal{E}(\widehat{f_{t}})$ is bounded above by $\rho \theta_{t}(\varphi_{t}^{*})^{2}$ on the event $\Omega_{t}$ by using some deterministic arguments. We next show that the event  $\Omega_{t}$ has a large probability and find a precise bound for $\phi_{t}^{*}$ in terms of the sample size and the Rademacher complexity using a peeling argument and applying Talagrand's concentration inequality \citep{ledoux1991probability}.

\textbf{Remark:} Theorem \ref{T4} provides an estimation error bound for the RNNs estimator obtained through ERM. The bound is established based on the generalization bound for the local Rademacher complexity given in Theorem \ref{T3}, and it can be sharper than the risk bound based on the global Rademacher complexity \citep{alquier2019}. Theorem \ref{T4} shows when the excess risk satisfies the Bernstein condition given in Condition \ref{A4}, the Frobenius norm distance between the RNN estimator $\widehat{f_ {t}}$ and the optimizer $f_ {t}^{*}$ in definition (\ref{fstar}) can be well-controlled by $\varphi_{t}^{*}$, and the excess risk of this estimator can also be well-controlled by $(\varphi_{t}^{*})^{2}$, indicating that the estimator converges to the theoretically optimal solution of the excess risk with a high probability. Section 8.2 in \citep{alquier2019} has discussed that the hinge loss satisfies the Bernstein condition. We show in Appendix \ref{S8-6} that the cross entropy loss satisfies the Bernstein condition given in Condition \ref{A4}.

In this section, we have established a sharp estimation error bound for the RNNs-based estimator obtained via ERM. In the next section, we will perform experiments on three public datasets to demonstrate our theoretical results.

\section{Experimental results}\label{S3}

In this section, we demonstrate the empirical performance of our established Rademacher complexity bound and compare it with the ones given in \cite{tu2019understanding}, \cite{zhang2018stabilizing} and \cite{chen2019}. We first introduce three real datasets in subsection \ref{S3-1} and experimental settings in subsection \ref{S3-2}. Then, we provide a detailed comparison of numerical results on these datasets in  subsections 
\ref{sec:comparision} and \ref{S3-4}.

\subsection{Real datasets}\label{S3-1}

We train vanilla RNNs on three real datasets to show the empirical validation of the derived Rademacher complexity bounds. The three datasets are described as follows: 
\begin{itemize}
    \item The PTB (Penn Tree Bank) \citep{marcinkiewicz1994building} dataset, commonly used for word-level prediction tasks, corresponds to the articles of the Wall Street Journal (WSJ). It contains 929,000 training words, 73,000 validation words, and 82,000 testing words with a vocabulary size of 10,000. This dataset is one of the most widely known and used corpus for the evaluation of models in sequence labeling, and it is available at \url{https://paperswithcode.com/dataset/penn-treebank}.
    \item The SST (Stanford Sentiment Treebank) \citep{socher2013recursive} dataset is a corpus containing fully labeled parse trees that allows for a comprehensive analysis of the compositional effects of sentiment in language. It was parsed with the Stanford parser and included a total of 215,154 unique phrases from those parse trees, each annotated by 3 human judges. Each phrase is categorized as either negative, somewhat negative, neutral, somewhat positive, or positive. This dataset is available at \url{https://paperswithcode.com/dataset/sst}.
    \item The SNLI dataset (Stanford Natural Language Inference) \citep{bowman2015large} consists of 570k sentence-pairs that have been manually labeled as ``entailment", ``contradiction", ``neutral" or ``-" where ``-" indicates that no agreement was reached. Annotators were instructed to judge the relationship between sentences given that they describe the same event. We randomly selected a subset of the data for training to reduce the computational costs. This dataset can be accessed at \url{https://paperswithcode.com/dataset/snli}.
\end{itemize}

\subsection{Experiment settings}\label{S3-2}
We use $\sigma_{h}=\text{ReLU}$ and $\sigma_{h}=\tanh$ as the activation functions of the hidden layers, respectively, to calculate the Rademacher complexity of the vanilla RNNs. The $\text{ReLU}$ function is unbounded, while the $\tanh$ function is bounded and it satisfies the assumption on $\sigma_{h}$ given in \cite{chen2019}.  
We set the hidden state $d_h \in \{32, 64, 128\}$ and the input $d_x \in \{30, 50, 100\}$. Accordingly, we train a 1-layered RNN model on word sequence length $t=\{10, 20, 30, 40\}$. We use the ramp loss with $\gamma=1$ in our data analysis.  We apply the source code which is available at \url{https://github.com/pytorch/examples/tree/main/word_language_model} to train the vanilla RNNs, and modify the code according to our setting. We use the default set-up for the hyperparameters in the source code with the dropout rate of 0.5, including setting the batch size and epochs to 20, and using the initial learning rate of 0.1 and the gradient clipping of 0.25 to prevent gradient exploding.  

\subsection{Comparison of our complexity bound with other bounds}\label{sec:comparision}

We show the complexity bounds of the vanilla RNNs given in Theorem 1 of \cite{tu2019understanding} (Bound 1), Theorem 5 of \cite{zhang2018stabilizing} (Bound 2),  Theorem 3 (Equation (3)) of \cite{chen2019} (Bound 3),  Equation (\ref{E16}) of Theorem \ref{T1} (Bound 4) and Equation (\ref{E20}) of Corollary \ref{L9} (Bound 4*). 
\begin{itemize}
  \item Bound 1: $B_{x,1}B_{W,1}B_{V,1}\Lambda/n$;
  \item Bound 2: $Btd\sqrt{\log(d)}\max\{(\rho_{h}M_{U})^{t-1},1\}M_{V}M_{W}\sqrt{t^2B_{U}^{2}+\frac{B_{W}^{2}}{M_{W}^{2}}+\frac{B_{V}^{2}}{M_{V}^{2}}}/\sqrt{n}$;
  \item Bound 3: $BM_{W}M_{U}\text{min}\{b\sqrt{d^{\prime}}, BM_{W}c_{t}^{\prime}\}c_{t}^{\prime}S_{F}\sqrt{d^{\prime}\log(d^{\prime})}/\sqrt{n}$;
  \item Bound 4 (Ours):   $g_{t}^{\prime}d\sqrt{\log(d)}\log\left(\rho_{h}nB_{x}B_{V}B_{W}c_{t}^{\prime}\right)/n$; \\
  \item \textcolor{black}{Bound 4* (Ours):  
   $g_{t}^{*}d\sqrt{\log(d)}\log\left(nB_{V}\min\{b\sqrt{nd},\rho_{h}B_{x}B_{W}c_{t}^{\prime}\}\right)/n$;}
\end{itemize}
where $d=\max\{d_{x},d_{h},d_{y}\}$, $d^{\prime}=\sqrt{d_{x}d_{h}+d_{h}^{2}+d_{h}d_{y}}$,
\begin{equation}\nonumber
\begin{split}
    b_{t}^{\prime }&=\frac{t(\rho _{h}M_{U})^{t-1}-c_{t}^{\prime }}{\rho
    _{h}M_{U}-1},c_{t}^{\prime }=\frac{(\rho _{h}M_{U})^{t}-1}{\rho _{h}M_{U}-1},\\
    \Lambda&=\frac{1}{1-\rho _{h}B_{U,1}}\left(\frac{1-\rho _{h}B_{U,1}^{t}}{1-\rho _{h}B_{U,1}}-t(\rho _{h}B_{U,1})^{t}\right),
    g_{t}^{\prime}=\rho _{h}B_{x}B_{V}B_{W}\max\{c_{t}^{\prime},\rho_{h}B_{U}b_{t}^{\prime}\},\\
    g_{t}^{*}&=B_{V}\min\{b\sqrt{nd},\rho _{h}B_{x}B_{W}c_{t}^{\prime}\}\max\{1,\rho_{h}B_{U}b_{t}^{\prime}/c_{t}^{\prime}\},
\end{split}
\end{equation}
in which $B_{x,1}$, $B_{U,1}$, $B_{W,1}$ and $B_{V,1}$ are the 1-norm of matrices $X$, $U$, $W$ and $V$, respectively; $B$ is the maximum of the row vector norms of the matrix $X$;
$M_{V}$ and $M_{W}$ are the spectral norm of matrices $V$ and $W$, and $S_{F}=B_{U}+B_{W}+B_{V}$.

\textbf{Remark:}  Bound 3 is provided in Theorem 3 (Equation (3)) of \cite{chen2019} when the hidden layer activation function $\sigma_h$ is bounded by a constant $b$ that can be satisfied by the $\tanh$ function. However, the $\text{ReLU}$ function is unbounded, so the complexity given in Theorem 3 of \cite{chen2019} can not be directly applied to this activation function. Similarly, Bound 4 is applied to the $\text{ReLU}$ function, while Bound 4* is tighter and can be used for the $\tanh$ function. 

Bound 2, Bound 3, and Bound 4* are based on the same assumptions of the spectral norm and Frobenius norm of the weight matrices, while Bound 1 is based on matrix 1-norm. Bound 1, Bound 2, and Bound 4* are provided without imposing a boundedness assumption on the activation function of the hidden layer, while Bound 3 and Bound 4* are assumed to have a natural upper bound $b$, which is satisfied by the $\tanh$ function not the ReLU function. 

In terms of the sample size $n$, Bound 2 and Bound 3  are the order of $O\left(1/\sqrt{n}\right)$, while Bound 1 and Bound 4 are the order of $O\left(1/n\right)$ and $O\left(\log(n)/n\right)$, respectively. However, Bound 1 is based on the 1-norm of $U$, which is often much larger than the spectral norm of $U$ involved in Bound 4. Moreover, we see that Bound 3 has one more $M_{U}^{t}$ term involved compared to Bound 4*. The spectral norm $M_U$ is often greater than one, so our Bound 4* should be tighter than Bound 2, especially when $t$ is large.

\subsection{Numerical results for comparison of different bounds}\label{S3-4}

The Rademacher complexities for the $\text{ReLU}$ and the $\tanh$ activation functions based on the PTB dataset are shown in Tables \ref{bb1} and \ref{bb2}, respectively. The Lipschitz parameters $\rho_{h}$ of both ReLU and $\tanh$ equal to 1.
In addition, Table \ref{bb2} also includes the experimental results for the datasets of SST and SNLI. 

\begin{table*}[!ht]
    \centering
    \caption{The complexity bounds for the $\text{ReLU}$ activation function for the PTB dataset. Imp\_perI (I=1,2) shows the improved percentage of Bound 4 (ours) compared to Bound I,  calculated by Imp\_perI=(Bound I-Bound 4)/Bound 4. The average improved percentage (Average) is calculated by averaging the improved percentages of the 12 cases.  
    }\label{bb1}
    \resizebox*{125mm}{!}{
    \begin{tabular}{ccccccccc}
    \hline
        $t$ & $n$ & $d_{x}$ & $d_h$ & Bound 1  & Bound 2 & Bound 4 (Ours) & Imp\_per1 & Imp\_per2 \\ \hline
        \multirow{3}{*}{10} & \multirow{3}{*}{92958} & 30 & 32   & 12.95 & 13.54 & \textbf{12.89} & 0.47\% & 5.04\%\\
          &  & 50 & 64  & 15.96 & 16.07 & \textbf{15.48} & 3.10\%	& 3.81\%
\\
          &  & 100 & 128  & 17.04 & 15.63 & \textbf{15.32} & 11.23\%	& 2.02\%
\\\hline
        \multirow{3}{*}{20} & \multirow{3}{*}{46479} & 30 & 32  & 25.78 & 21.43 & \textbf{20.57} & 25.33\%	& 4.18\%
\\
        &  & 50 & 64 & 24.07 & 20.45   & \textbf{19.93} & 20.77\%	& 2.61\%
\\
        &  & 100 & 128 & 28.26 & 22.60   & \textbf{22.28} & 26.84\%	 & 1.44\%
\\\hline
        \multirow{3}{*}{30} & \multirow{3}{*}{30986} & 30 & 32 & 23.51 & 15.22   & \textbf{14.57} & 61.36\%	& 4.46\% \\
        &  & 50 & 64 & 27.30 & 14.75  & \textbf{14.34} & 90.38\% & 2.86\%
\\
        &  & 100 & 128 & 36.22 & 23.15  & \textbf{22.89} & 58.24\%	& 1.14\%
\\\hline
        \multirow{3}{*}{40} & \multirow{3}{*}{23239} & 30 & 32 & 28.90 & 15.98  & \textbf{15.28} & 89.14\%	& 4.58\% \\
        &  & 50 & 64 & 34.34 & 15.38  & \textbf{14.93} & 130.01\%	& 3.01\%
\\
        &  & 100 & 128 & 39.84 & 17.66  & \textbf{17.49} & 127.79\%	 & 0.97\%
\\\hline
        &  &  Average &  &  &  &  & 53.72\% & 3.01\% \\\hline
    \end{tabular}
    }
\end{table*}

\begin{table*}[!ht]
\centering
\caption{The complexity bounds for the $\tanh$ activation function. Imp\_perI denotes the improved percentage of Bound 4* (ours) compared to Bound I for I=2,3. The average improved percentage (Average) is calculated by averaging the improved percentages of the 18 cases.}\label{bb2}
\resizebox*{135mm}{!}{
\begin{tabular}{ccccccccccc}
\hline
    Datasets & $t$ & $n$ & $d_{x}$ & $d_h$ & Bound 1 & Bound 2 & Bound 3 & Bound 4*  (Ours)  & Imp\_per2 & Imp\_per3 \\ \hline
     \multirow{12}{*}{PTB} & \multirow{3}{*}{10} & \multirow{3}{*}{92958} & 30 & 32 & 11.78 & 12.30  & 9.23  & \textbf{8.42} & 46.08\% & 9.62\%\\
     & &  & 50 & 64 & 13.23 & 12.71  & 9.75  & \textbf{8.71} & 45.92\% & 11.94\% \\
     & &  & 100 & 128 & 14.52 & 13.03 & 10.31  & \textbf{8.97} & 45.26\% &  14.94\% \\ \cline{2-11}
     & \multirow{3}{*}{20} & \multirow{3}{*}{46479} & 30 & 32 & 17.55 & 13.42  & 9.91  & \textbf{8.79} & 52.67\% & 12.74\% \\
     & &  & 50 & 64 & 19.92 & 13.52 & 10.08  & \textbf{9.12} & 48.25\% & 10.53\%\\
     & &  & 100 & 128 & 22.70 & 14.02 & 10.63  & \textbf{9.42} & 48.83\% & 12.85\% \\ \cline{2-11}
     & \multirow{3}{*}{30} & \multirow{3}{*}{30986} & 30 & 32 & 21.23 & 13.44 & 9.63  & \textbf{9.04} & 48.67\% & 6.53\%\\
     & &  & 50 & 64 & 26.38 & 13.64 & 9.95  & \textbf{9.37} & 45.57\% & 6.19\%\\
     & &  & 100 & 128 & 30.79 & 14.77 & 11.19  & \textbf{9.66} & 52.90\% & 15.84\%\\ \cline{2-11}
     & \multirow{3}{*}{40} & \multirow{3}{*}{23239} & 30 & 32 & 26.66 & 14.29 & 10.22  & \textbf{9.21} & 55.16\% & 10.97\% \\
     & &  & 50 & 64 & 32.67 & 13.76 & 9.81  & \textbf{9.54} & 44.23\% & 2.83\% \\
     & &  & 100 & 128 & 37.67 & 14.81 & 11.03  & \textbf{9.83} & 50.66\% & 12.21\% \\\hline
     \multirow{3}{*}{SST} & \multirow{3}{*}{20} & \multirow{3}{*}{24534} & 30 & 32 & 25.67 & 15.80 & 12.31 & \textbf{10.14} & 55.82\% & 21.40\%\\ 
     & & & 50 & 64 & 26.18 & 15.39 & 12.20 & \textbf{10.18} & 51.18\% & 19.84\%\\ 
     & & & 100 & 128 & 26.87 & 15.53 & 12.22 & \textbf{10.31} & 50.63\% & 18.53\%\\ \hline
     \multirow{3}{*}{SNLI} & \multirow{3}{*}{20} & \multirow{3}{*}{41759} & 30 & 32 & 24.63 & 14.51 & 11.78 & \textbf{9.60} & 51.15\% & 22.71\%\\ 
     & & & 50 & 64 & 24.97 & 15.69 & 12.08 & \textbf{10.26} & 52.92\% &17.74\%\\ 
     & & & 100 & 128 & 26.41 & 15.75 & 12.12 & \textbf{10.30} & 52.91\% & 17.67\% \\ \hline
     & & Average & & & & & & & 49.93\% & 13.80\% \\ \hline
\end{tabular}
}
\end{table*}

Our empirical results align with our theoretical bounds. First, we observe that our bound is the tightest for all cases in both tables. We calculate the improved percentage of our bound (Bound 4) compared to the existing ones by using the formula:  Imp\_perI=(Bound I-Bound 4)/Bound 4, for Bound I=1,2,3. Tables \ref{bb1} and \ref{bb2}  show that our bound has a decrease by an average of 3.01\% compared to Bound 2 (the second tightest bound in Table \ref{bb1}) when the ReLU activation function is used;  it has a significant decrease by an average of 13.80\% compared to Bound 3 (the second tightest bound in Table \ref{bb2}) when the $\tanh$ activation function is used.

Table \ref{bb1} shows that when the $\text{ReLU}$ activation function is applied in the hidden layers, our bound (Bound 4) given in Theorem \ref{T1} is tighter than Bound 1 and Bound 2. It corroborates with our theoretical findings as aforementioned. Bound 3 is not included in Table \ref{bb1}, because the ReLU function does not satisfy the boundedness assumption required by Bound 3. When the dimensions $d_{x}$ and $d_h$ are larger or the sequence length $t$ is larger,  Bound 4 becomes tighter than Bound 1. This can be explained by the fact that the spectral norm is often smaller than the 1-norm of the weight matrices. We see that our bound (Bound 4) significantly improves Bound 1 by an average of 53.72\%, and it improves Bound 2 by an average of 3.01\% in Table \ref{bb1}. When the $\tanh$ function which satisfies the boundedness assumption is used, we can obtain an even tighter bound given in Bound 4*, which significantly improves Bound 2 by an average of 49.93\% as shown in Table \ref{bb2}.

Table \ref{bb2} shows the experimental results for the three real datasets PTB, SST and SNL1 when the $\tanh$ activation function is used. 
 Our Bound 4* is the tightest among the four bounds. Since the pattern of four bounds is similar among the three datasets for given $t$ and $n$ values, we only report the result for $t=20$ for the datasets of SST and SNLI to save space. Bound 3 becomes the second tightest, but it is still larger than Bound 4* (ours). Bound 4* improves Bound 3 by an average of $13.80\%$. The improvement of Bound 4* over the other three bounds becomes more significant when the dimension or the sequence length becomes larger.

Last but not least, Theorems \ref{T1} and \ref{T2} show that our theoretical generalization bounds depend on the dimensions of ($d_h$ and $d_x$) the hidden state and input data in a polynomial order, implying that RNNs do not suffer significantly from the dimensionality problem. This is also corroborated by our numerical results in which we find that RNNs are not sensitive to the choice of $d_h$. In Tables \ref{bb1} and \ref{bb2}, we see that when $t$ and $n$ are fixed, the bounds are moderately increased from $d_h=32$ to $d_h=128$. Furthermore, we observe that the bounds for the $\tanh$ (bounded) activation function shown in Table \ref{bb2} are tighter and less severely affected by the sequence length $t$ than the bounds for the ReLU (unbounded) activation function given in Table \ref{bb1}, confirming our theoretical results given in Theorems  \ref{T1} and \ref{T2}.

\section{Discussion}\label{S4}

This paper establishes a new generalization bound for vanilla RNNs under the PAC-learning framework, and it can be applied to a variety of loss functions. When the ramp loss is used as in \cite{zhang2018stabilizing} and \cite{chen2019}, we show that our bound is tighter than theirs under the same assumption on the Frobenius
and spectral norms of the weight matrices and also tighter than the bound given in \cite{tu2019understanding} in different scenarios. We also derive a generalization bound that only involves the Frobenius norms, and it meets the computational needs in real data applications.  When the activation operator is bounded as assumed in \cite{chen2019}, we further obtain a tighter generalization bound, and show that our bound involves one less $M_{U}^{t}$ than theirs. Moreover, we derive a sharp estimation error bound for the RNN-based estimator obtained through ERM when the loss function satisfies a Bernstein condition. The established properties provide an important theoretical basis and guidance for the statistical analysis of real data using RNNs. {\color{black}Numerically, we demonstrate the empirical performance of our established generalization bounds using three public datasets and show that our bound is tighter than the existing ones in different scenarios. For future work, we will conduct more experimental studies to evaluate the properties of RNN estimators, including studying the biases in the evaluations, the robustness of the RNN estimators in the selection of hyperparameter values, and the optimization techniques for predictions and classifications.}

\subsection{Other sequence models related to RNNs}

RNNs, firstly introducing memory units, are the foundational models in sequence-to-sequence learning tasks, so understanding their statistical learning properties is important for sequential data analysis.  This paper contributes by bridging the gap between theoretical understanding and practical applications of RNNs. The theoretical framework developed in our paper can be extended possibly to the variants of RNNs, including Long Short-Term Memories (LSTM) \citep{hochreiter1997long}, Gated Recurrent Units (GRU) \citep{cho2014learning}, and bidirectional RNNs. To be specific, LSTM introduces three gates (the input gate, the forget gate, and the output gate) and GRU has two gates (the reset gate and the update gate) to alleviate the long-term memory dependency of vanilla RNNs, respectively. The impact of gating on the empirical Rademacher complexity needs to be carefully investigated in the potential extensions. Bidirectional RNNs process input sequences in both forward and backward directions in each time step to better capture dynamic features between sequences, so they involve two distinct (feedforward and backward) recurrent hidden layers. To extend our generalization bounds for bidirectional RNNs, we need to calculate the model complexity of the two hidden layers. 

To facilitate parallel computing and alleviate the long-term memory dependency problem of RNNs, \cite{vaswani2017attention} proposed the Transformer model. Recently, xLSTM \citep{beck2024xlstm} re-builds the memory units of RNNs and introduces exponential gating and parallelizable matrix memory structures to mitigate some limitations of RNNs. The theoretical properties in these more complicated models need further and careful investigation. We will extend our learning theory in these models in future work.

\newpage
 \begin{appendices} \label{S8}

\setcounter{equation}{0}
\renewcommand\theequation{A.\arabic{equation}}

\setcounter{lemma}{0} 
\renewcommand{\thelemma}{A.\arabic{lemma}}

In this appendix, we provide the proofs of the generalization error bound with respect to RNNs in Appendix \ref{S8-1} and the estimation error bound of RNNs in Appendix \ref{S8-2-3}.

\section{Proof of the generalization error bound} \label{S8-1}

In this section, we provide the technical proofs for  Lemma \ref{L88}, Theorem \ref{T1} and Corollary \ref{L9}. Lemmas \ref{L2}-\ref{L7} given in subsection \ref{S8-1-6} contain some intermediate results  used to prove Theorem \ref{T1}.

\subsection{Proof of Lemma \ref{L88}} \label{S8-1-5}
Under Assumption \ref{A5} (i) where $\ell_{f_{t}}$ is bounded in $C_{t}$, it follows by standard properties of Rademacher complexity (Theorem 3.3 in Chapter 3, \citep{mohri2018}) that with probability at least $1-\delta$, for each $f_{t} \in \mathcal{F}_{t}$ satisfies 
\begin{equation*}
    \mathcal{R}(f_{t}) \leq \widehat{\mathcal{R}}(f_{t})+2\mathfrak{R}_{S}(\mathcal{F}_{\ell,t})+3C_{t}\sqrt{\frac{\log(2/\delta)}{2n}}.
\end{equation*}
The left side of the bound in Lemma \ref{L88} follows $\mathds{P}[ \hat{z}_{t} \neq z_{t}] \leq \mathcal{R}(f_{t})$ by definition.

\subsection{Proof of Theorem \ref{T1}} \label{S8-1-6}

In the following lemma, we first derive the upper bound $\|h_{t}\|_{F}$. 
\begin{lemma}\label{L2}
    (a) Under Assumptions \ref{A1}-\ref{A3}, for $t \geq 0$, the Frobenius norm of $h_{t}$ is bounded by
    $$\|h_{t}\|_{F} \leq \rho_{h}B_{W}B_{x}c_{t},$$
    where $c_{t}=\frac{(\rho_{h}B_{U})^{t}-1}{\rho_{h}B_{U}-1}$.

  \textbf{Proof.} We prove this by induction. Observe that for $t \geq 1$, we have
    \begin{equation*}
        \begin{split}
            \|h_{t}\|_{F}& = \|\sigma_{h}(Uh_{t-1}+Wx_{t})\|_{F}\\
            & = \|\sigma_{h}(Uh_{t-1}+Wx_{t})-\sigma_{h}(0)\|_{F}\\
            & \leq \rho_{h}(\|Uh_{t-1}\|_{F}+\|Wx_{t}\|_{F})\\
            & \leq \rho_{h}\|Uh_{t-1}\|_{F}+\rho_{h}\|Wx_{t}\|_{F}\\
            & \leq \rho_{h}B_{U}\|h_{t-1}\|_{F}+\rho_{h}B_{W}B_{x}\\
            & \leq (\rho_{h}B_{U})^{t}\|h_{0}\|_{F}+\rho_{h}B_{W}B_{x}\sum_{j=0}^{t-1}(\rho_{h}B_{U})^{j}\\
            & = \rho_{h}B_{W}B_{x}\frac{(\rho_{h}B_{U})^{t}-1}{\rho_{h}B_{U}-1}= \rho_{h}B_{W}B_{x}c_{t}.\\
        \end{split}
    \end{equation*}
    Clearly, $ \|h_{0}\|_{F}=0$ satisfies the upper bound. The remaining task is to bound $\|h_{t}-h_{t}^{\prime}\|_{F}$ in terms of the Frobenius norms of the difference of weight matrices, $\|U-U^{\prime}\|_{F} $ and $\|W-W^{\prime}\|_{F}$.

    (b) Under Assumptions \ref{A1}-\ref{A3} and the spectral norm of $U$ can be further bounded by $M_{U}$, for $t \geq 0$, the Frobenius norm of $h_{t}$ is bounded by
    $$\|h_{t}\|_{F} \leq \rho_{h}B_{W}B_{x}c_{t}^{\prime},$$
    where $c_{t}^{\prime}=\frac{(\rho_{h}M_{U})^{t}-1}{\rho_{h}M_{U}-1}$.

  \textbf{Proof.} We prove by induction. Observe that for $t \geq 1$, we have
    \begin{equation*}
        \|h_{t}\|_{F} 
        \leq \rho_{h}\|Uh_{t-1}\|_{F}+\rho_{h}\|Wx_{t}\|_{F}
        \leq \rho_{h}M_{U}\|h_{t-1}\|_{F}+\rho_{h}B_{W}B_{x}.
    \end{equation*}
Following the same procedures of Lemma \ref{L2} (a) recursively, we have 
$\|h_{t}\|_{F} = \rho_{h}B_{W}B_{x}\frac{(\rho_{h}M_{U})^{t}-1}{\rho_{h}M_{U}-1}
               = \rho_{h}B_{W}B_{x}c_{t}^{\prime}$.
\end{lemma}

\textcolor{black}{\textbf{Remark: }Deriving the bound in part (b) of Lemma \ref{L2} shares a similar proof procedure as that in part (a). The only difference is that we replace $B_{U}$ by $M_{U}$ in part (b) in the bound for the Frobenius norm $\|Uh_{t-1}\|_{F}$ when an additional condition that the spectral norm of $U$ is bounded by $M_{U}$ is assumed. The rest proof in part (b) follows the same procedure as in part (a). As a result, the bound in part (b) is tighter than the one given in part (a) under this additional assumption. }

We now calculate an intermediate result which will be used by Lemma \ref{L3}.
\begin{equation} \label{L4}
    \begin{split}
    \sum_{j=0}^{t-1}\frac{(\rho_{h}B_{U})^{t-1}-(\rho_{h}B_{U})^{t-1-j}}{\rho_{h}B_{U}-1} &= \sum_{j=0}^{t-2}[(j+1)(\rho_{h}B_{U})^{j}]\\
    &=\frac{t(\rho_{h}B_{U})^{t-1}-c_{t}}{\rho_{h}B_{U}-1}
    =b_{t}. \\
    \end{split}
\end{equation}

Next, we need to bound the differences of $h_{t}-h_{t}^{\prime}$ in the following lemma.

\begin{lemma}\label{L3}
    (a)
    Under Assumptions \ref{A1}-\ref{A3}, for $t \geq 1$, the difference of hidden states  $h_{t}$ and $h_{t}^{\prime}$ satisfies
    $$\|h_{t}-h_{t}^{\prime}\|_{F} \leq \rho_{h}^2B_{W}B_{x}\|U-U^{\prime}\|_{F}b_{t}+\rho_{h}B_{x}\|W-W^{\prime}\|_{F}c_{t},$$
    where $L_{U,t}= a\rho_{h}B_{W}B_{V}b_{t}$ and $L_{W,t}=aB_{V}c_{t}$ with $a=\rho_{h}B_{x}$, $b_{t}=\frac{t(\rho_{h}B_{U})^{t-1}-c_{t}}{\rho_{h}B_{U}-1}$ and $c_{t}=\frac{(\rho_{h}B_{U})^{t}-1}{\rho_{h}B_{U}-1}$.

    \textbf{Proof.} Similar to the proof of Lemma \ref{L2}, we use induction.
    \begin{equation*}
        \begin{split}
            & \quad \|h_{t}-h_{t}^{\prime}\|_{F}  = \|\sigma_{h}(Uh_{t-1}+Wx_{t})-\sigma_{h}(U^{\prime}h_{t-1}^{\prime}+W^{\prime}x_{t})\|_{F}\\
            & \leq \rho_{h}\|(Uh_{t-1}+Wx_{t})-(U^{\prime}h_{t-1}^{\prime}+W^{\prime}x_{t})\|_{F}\\
            & \leq \rho_{h}\|(Uh_{t-1}-U^{\prime}h_{t-1})+(U^{\prime}h_{t-1}-U^{\prime}h_{t-1}^{\prime})
            +(Wx_{t}-W^{\prime}x_{t})\|_{F}\\
            & \leq \rho_{h}\|Uh_{t-1}-U^{\prime}h_{t-1}\|_{F}
             +\rho_{h}\|U^{\prime}h_{t-1}-U^{\prime}h_{t-1}^{\prime}\|_{F} +\rho_{h}\|Wx_{t}-W^{\prime}x_{t}\|_{F}\\
            & \leq \rho_{h}\|U-U^{\prime}\|_{F}\|h_{t-1}\|_{F}+\rho_{h}B_{U}\|h_{t-1}-h_{t-1}^{\prime}\|_{F} +\rho_{h}B_{x}\|W-W^{\prime}\|_{F}\\
            & \leq (\rho_{h}B_{U})^{t}\|h_{0}-h_{0}^{\prime}\|_{F}
             +\rho_{h}\|U-U^{\prime}\|_{F}\sum_{j=0}^{t-1}(\rho_{h}B_{U})^{t-1-j}\|h_{j}\|_{F}\\
            & \quad +\rho_{h}B_{x}\|W-W^{\prime}\|_{F}\sum_{j=0}^{t-1}(\rho_{h}B_{U})^{j}\\
            & \leq \rho_{h}\|U-U^{\prime}\|_{F}\sum_{j=0}^{t-1}(\rho_{h}B_{U})^{t-1-j}\rho_{h}B_{W}B_{x}\frac{(\rho_{h}B_{U})^{j}-1}{\rho_{h}B_{U}-1} + \rho_{h}B_{x}\|W-W^{\prime}\|_{F}c_{t}\\
            &=\rho_{h}^2B_{W}B_{x}\|U-U^{\prime}\|_{F}\sum_{j=0}^{t-1}\frac{(\rho_{h}B_{U})^{t-1}-(\rho_{h}B_{U})^{t-1-j}}{\rho_{h}B_{U}-1} + \rho_{h}B_{x}\|W-W^{\prime}\|_{F}c_{t} \\
            & = \rho_{h}^2B_{W}B_{x}\|U-U^{\prime}\|_{F}\frac{t(\rho_{h}B_{U})^{t-1}-c_{t}}{\rho_{h}B_{U}-1} +\rho_{h}B_{x}\|W-W^{\prime}\|_{F}c_{t}\\
            & = \rho_{h}^2B_{W}B_{x}\|U-U^{\prime}\|_{F}b_{t}+\rho_{h}B_{x}\|W-W^{\prime}\|_{F}c_{t},\\
        \end{split}
    \end{equation*}
    where the last inequality holds by Equation (\ref{L4}).

    (b) Under Assumptions \ref{A1}-\ref{A3} and the spectral norm of $U$ can be further bounded by $M_{U}$, for $t \geq 1$, the difference of hidden states  $h_{t}$ and $h_{t}^{\prime}$ satisfies
    $$\|h_{t}-h_{t}^{\prime}\|_{F} \leq \rho_{h}^2B_{W}B_{x}\|U-U^{\prime}\|_{F}b_{t}^{\prime}+\rho_{h}B_{x}\|W-W^{\prime}\|_{F}c_{t}^{\prime},$$
    where $L_{U,t}= a\rho_{h}B_{W}B_{V}b_{t}^{\prime}$ and $L_{W,t}=aB_{V}c_{t}^{\prime}$ with $a=\rho_{h}B_{x}$, $b_{t}^{\prime}=\frac{t(\rho_{h}M_{U})^{t-1}-c_{t}^{\prime}}{\rho_{h}M_{U}-1}$ and $c_{t}^{\prime}=\frac{(\rho_{h}M_{U})^{t}-1}{\rho_{h}M_{U}-1}$.

    The proof of Lemma \ref{L3} (b) can follow by Lemma \ref{L2} (b) and Lemma \ref{L3} (a), so we omit the details.
\end{lemma}

In the following lemma, we characterize the Lipschitz property of $\|y_{t}-y_{t}^{\prime}\|_{F}$ with regard to the model parameters.

\begin{lemma}\label{L1}
    Under Assumptions \ref{A1}-\ref{A3}, given the input $\{x_{1},...,x_{t}\}$, $\|y_{t}\|_{F}$ is Lipschitz in $U, V$ and $W$, i.e.,
    \begin{equation*}
        \begin{split}
        \|y_{t}-y_{t}^{\prime}\|_{F} \leq  L_{V,t}\|V-V^{\prime}\|_{F}+L_{U,t}\|U-U^{\prime}\|_{F}+L_{W,t}\|W-W^{\prime}\|_{F},
        \end{split}
    \end{equation*}
    where (a) $L_{V,t}=aB_{W}c_{t}$, $L_{U,t}= a\rho_{h}B_{W}B_{V}b_{t}$ and $L_{W,t}=aB_{V}c_{t}$; (b) If $\|U\|_{\sigma} \leq M_{U}$, $L_{V,t}=aB_{W}c_{t}^{\prime}$, $L_{U,t}= a\rho_{h}B_{W}B_{V}b_{t}^{\prime}$ and $L_{W,t}=aB_{V}c_{t}^{\prime}$.
\end{lemma}
Note that Lemma \ref{L1} is the key to the proof of our generalization bound. According to Lemma \ref{L1}, 
we calculate the empirical Rademacher complexity $\mathfrak{R}_{S}(\mathcal{F}_{\ell,t})$ in the Theorem \ref{T1} and the Corollary \ref{L9}.

\textbf{Proof.} This lemma begins with matrix Frobenius norms. Given the weight matrices $U, V, W$ and $U^{\prime}, V^{\prime}, W^{\prime}$, consider the $t$-th outputs $y_{t}$ and $y_{t}^{\prime}$ of vanilla RNNs,
    \begin{equation*}
        \begin{split}
            & \quad \|y_{t}-y_{t}^{\prime}\|_{F} 
               =\|Vh_{t}-V^{\prime}h_{t}^{\prime}\|_{F} \\
             & = \|Vh_{t}-V^{\prime}h_{t}+V^{\prime}h_{t}-V^{\prime}h_{t}^{\prime}\|_{F} \\
             & \leq \|Vh_{t}-V^{\prime}h_{t}\|_{F} +\|V^{\prime}h_{t}-V^{\prime}h_{t}^{\prime}\|_{F}\\
             & \leq \|V-V^{\prime}\|_{F}\|h_{t}\|_{F} +B_{V}\|h_{t}-h_{t}^{\prime}\|_{F}\\
             & \leq \rho_{h}B_{W}B_{x}c_{t}\|V-V^{\prime}\|_{F}
              +\rho_{h}^2B_{W}B_{V}B_{x}b_{t}\|U-U^{\prime}\|_{F} +\rho_{h}B_{V}B_{x}c_{t}\|W-W^{\prime}\|_{F} \\
             & \leq aB_{W}c_{t}\|V-V^{\prime}\|_{F}+a\rho_{h}B_{W}B_{V}b_{t}\|U-U^{\prime}\|_{F}
              +aB_{V}c_{t}\|W-W^{\prime}\|_{F} \\
             & =L_{V,t}\|V-V^{\prime}\|_{F}+L_{U,t}\|U-U^{\prime}\|_{F}
             +L_{W,t}\|W-W^{\prime}\|_{F},
        \end{split}
    \end{equation*}
where $a=\rho_{h}B_{x}$ and the third inequality holds by Lemma \ref{L2} and Lemma \ref{L3}.

Then, we can further obtain the covering numbers of weight matrices of $U$, $V$ and $W$ by the following lemma. 
\begin{lemma}(Lemma 3.2. in \cite{bartlett2017})\label{L5}
    Let $\mathcal{G}=\{A\in \mathbb{R}^{d_{1}\times d_{2}}:\|A\|_{F}\leq\lambda\}$ be the set of matrices with bounded Frobenius norm and  $\epsilon>0$ be given. The covering number
    $\log \mathcal{N}(\mathcal{G},\epsilon,\|\cdot\|_{F})$ is upper bounded by
    \begin{equation*}
       \log \mathcal{N}(\mathcal{G},\epsilon,\|\cdot\|_{F}) \leq \left \lceil \frac{\lambda^{2}d_{1}d_{2}}{\epsilon^{2}} \right \rceil \log(2d_{1}d_{2}),
    \end{equation*}
    where $\lceil \varsigma \rceil$ denotes the integer part of $\varsigma$.
\end{lemma}
By Lemma \ref{L5}, we can directly write out the upper bounds on the covering numbers of weight matrices,
\begin{equation}\nonumber
    \begin{split}
         &\log \mathcal{N}(U,\frac{\epsilon}{3L_{U,t}},\|\cdot\|_{F}) \leq  \frac{9 d_{h}^2B_{U}^{2}L_{U,t}^2}{\epsilon^{2}} \log(2d_{h}^{2}),\\
         &\log \mathcal{N}(V,\frac{\epsilon}{3L_{V,t}},\|\cdot\|_{F}) \leq \frac{9 d_{h}d_{y}B_{V}^{2}L_{V,t}^2}{\epsilon^{2}} \log(2d_{y}d_{h}), \\
         &\log \mathcal{N}(W,\frac{\epsilon}{3L_{W,t}},\|\cdot\|_{F}) \leq  \frac{9 d_{x}d_{h}B_{W}^{2}L_{W,t}^2}{\epsilon^{2}} \log(2d_{h}d_{x}).
    \end{split}
\end{equation}

Our goal is to construct a covering $\mathcal{C}(\mathcal{F}_{t},\epsilon,\|\cdot\|_{F})$, i.e., for any $f_{t}\in \mathcal{F}_{t}$, there exists $\widehat{f}_{t}\in \mathcal{F}_{t}$, for any input data $X_{t}$, satisfying
$$\underset{X_{t}}{\sup}\|f_{t}(X_{t})-\widehat{f}_{t}(X_{t})\|_{F} \leq \epsilon. $$
Note that $f_{t}$ is determined by weight matrices $U$, $V$ and $W$. By Lemma \ref{L1}, we have
\begin{equation}\nonumber
    \begin{split}
    \underset{X_{t}}{\sup}\|f_{t}(X_{t})-\widehat{f}_{t}(X_{t})\|_{F} \leq L_{V,t}\|V-\widehat{V}\|_{F}+L_{U,t}\|U-\widehat{U}\|_{F}
    +L_{W,t}\|W-\widehat{W}\|_{F}.
    \end{split}
\end{equation}
Then it is enough to construct three matrix coverings, $\mathcal{C}(U,\frac{\epsilon}{3L_{U,t}},\|\cdot\|_{F})$, $\mathcal{C}(V,\frac{\epsilon}{3L_{V,t}},\|\cdot\|_{F})$, and $\mathcal{C}(W,\frac{\epsilon}{3L_{W,t}}\|\cdot\|_{F})$. Their Cartesian product gives us the covering $\mathcal{C}(\mathcal{F}_{t},\epsilon,\|\cdot\|_{F})$.
\begin{equation}\label{L6}
    \begin{split}
      & \quad \log \mathcal{N}(\mathcal{F}_{t},\epsilon,\|\cdot\|_{F}) \\
      & \leq \frac{9d_{h}^{2}B_{U}^{2}L_{U,t}^2}{\epsilon^{2}} \log(2d_{h}^{2})
      +\frac{9d_{h}d_{y}B_{V}^{2}L_{U,t}^2}{\epsilon^{2}} \log(2d_{y}d_{h}) +\frac{9d_{x}d_{h}B_{W}^{2}L_{U,t}^2}{\epsilon^{2}} \log(2d_{h}d_{x})\\
    & = \frac{9d_{h}^{2}a^{2}\rho_{h}^2B_{U}^{2}B_{V}^{2}B_{W}^{2}b_{t}^2}{\epsilon^{2}} \log(2d_{h}^{2}) +\frac{9d_{h}d_{y}a^{2}B_{V}^{2}B_{W}^{2}c_{t}^2}{\epsilon^{2}} \log(2d_{y}d_{h})\\
     & \quad +\frac{9d_{x}d_{h}a^{2}B_{V}^{2}B_{W}^{2}c_{t}^2}{\epsilon^{2}} \log(2d_{h}d_{x})\\
    & \leq \frac{27d^{2}g_{t}^{2}}{\epsilon^{2}} \log(2d^{2}),\\
    \end{split}
\end{equation}
where $d=\max\{d_{x},d_{y},d_{h}\}$, $g_{t}=aB_{V}B_{W}\max\{c_{t},\rho_{h}B_{U}b_{t}\}$ and the first inequality holds by Lemma \ref{L5}.

Finally, using the following lemma, we can obtain the empirical Rademacher complexity of vanilla RNNs by the covering numbers of weight matrices.
\begin{lemma} (Lemma A.5. in \cite{bartlett2017} and Lemma 9. in \cite{chen2019}) \label{L7}
    Let $\mathcal{F}$ be a real-valued function class taking values in $[-r,r]$, and assume $0 \in \mathcal{F}$. Let $S=(s_{1},...,s_{n})$ be given points, then
    \begin{equation*}
        \mathfrak{R}_{S}(\mathcal{F}) \leq \underset{\alpha>0} {\inf} \left( \frac{4\alpha}{\sqrt{n}}+\frac{12}{n}\int_{\alpha}^{2r\sqrt{n}}\sqrt{\log\mathcal{N}(\mathcal{F},\epsilon,\|\cdot\|_{F})}\;d\epsilon\right).
    \end{equation*}
Lemma \ref{L7} is a slight variant of the standard Dudley entropy integral bound on the empirical Rademacher complexity and \cite{bartlett2017} provides a rigorous proof for this lemma (A.3 Dudley Entropy Integral on page 14).
    Taking $\mathcal{F}=\mathcal{F}_{t}$, we can easily verify that $\mathcal{F}_{t}$ takes values in $[-r,r]$ with $r=B_{V}\|h_{t}\|_{F}\leq aB_{V}B_{W}c_{t}$. Then by Lemma \ref{L7}, we have 
    \begin{equation*}
        \mathfrak{R}_{S}(\mathcal{F}_{t}) \leq \underset{\alpha>0} {\inf} \left( \frac{4\alpha}{\sqrt{n}}+\frac{12}{n}\int_{\alpha}^{2r\sqrt{n}}\sqrt{\log\mathcal{N}(\mathcal{F}_{t},\epsilon,\|\cdot\|_{F})}\;d\epsilon\right).
    \end{equation*}
    Then, we bound the integral as follows,
    \begin{equation*}
        \begin{split}
               \int_{\alpha}^{2r\sqrt{n}}\sqrt{\log\mathcal{N}(\mathcal{F}_{t},\epsilon,\|\cdot\|_{F})}\;d\epsilon\;
            & \leq \int_{\alpha}^{2r\sqrt{n}}\sqrt{\frac{27d^{2}g_{t}^{2}}{\epsilon^{2}} \log(2d^{2})}\;d\epsilon\\
            & = 3dg_{t}\sqrt{3\log(2d^{2})}\log\left(\frac{2r\sqrt{n}}{\alpha}\right),\\
        \end{split}
    \end{equation*}
   where the first inequality follows Inequation \ref{L6}. Picking $\alpha=\frac{1}{\sqrt{n}}$ is enough to give us an upper bound on $\mathfrak{R}_{S}(\mathcal{F}_{t})$,
    \begin{equation*}
        \mathfrak{R}_{S}(\mathcal{F}_{t}) \leq \frac{4}{n}+\frac{36dg_{t}}{n}\sqrt{3\log(2d^{2})}\log(2rn).
    \end{equation*}

Finally, by Talagrand's lemma \citep{mohri2018} and $\ell_{f_{t}}$ being $\rho$-Lipschitz, we have
    \begin{equation*}
        \mathfrak{R}_{S}(\mathcal{F}_{\ell,t}) \leq 2\rho\mathfrak{R}_{S}(\mathcal{F}_{t})
        \leq \frac{8\rho}{n}+\frac{72\rho dg_{t}}{n}\sqrt{3\log(2d^{2})}\log(2rn).
    \end{equation*}
\end{lemma}
Then the proof procedure of Theorem \ref{T1} has been completed. Combining Lemma \ref{L88}, We can immediately obtain the result of Theorem \ref{T2}.

\subsection{Proof of Corollary \ref{L9}.}

Since $\sigma_{h}$ is entry-wise bounded by 
$b$, $\|h_{t}\|_{F} \leq b\sqrt{nd}$. Combining Lemma \ref{L2} (b), we can immediately obtain $\|h_{t}\|_{F} \leq \min\{b\sqrt{nd},\rho _{h}B_{x}B_{W}c_{t}^{\prime}\}$. The following procedure is similar to calculating Rademacher complexity $\mathfrak{R}_{S}(\mathcal{F}_{\ell,t})$ in Theorem \ref{T1} without Assumption \ref{A7}, so we ignore the details.

\section{The upper bound and Lipschitz continuity of \texorpdfstring{$\ell_{f_{t}}$}{}} \label{S8-3}

\setcounter{lemma}{0} 
\renewcommand{\thelemma}{B.\arabic{lemma}}

\subsection{Proof of Lemma \ref{Loss2}} \label{S8-1-0}
Since $\ell_{f_{t}}$ is $\rho$-Lipschitz  with respect to $f_{t}$ and $\|f_{t}\|_{\infty} \leq \omega_{t}$, we have
\begin{equation}
    \begin{split}
    \|\ell(q(f_{t}(X_{t})),z_{t})\|_{\infty} & =\|\ell(q(f_{t}(X_{t})),z_{t})-\ell(0,z_{t})\|_{\infty} \\
    & \leq \rho\|f_{t}(X_{t})-f_{t}(X_{t}^{\prime})\|_{\infty}
      \leq 2\rho \omega_{t}.\\
    \end{split}
\end{equation}

\subsection{Proof of Lipschitz continuity of loss functions} \label{S8-1-10}

We show the Lipschitz continuity of the cross entropy loss $\ell_{f_{t}}$ in the following lemma.
\begin{lemma}\label{L0}
Let $f_{t,k}=f_{t,k}(X_{t})$, $q_{t,k}=\frac{e^{f_{t,k}}}{\sum_{k=1}^{K}e^{f_{t,k}}}$ and $\ell_{f_{t}}=\ell(q_{t},z_{t})=-\sum_{k=1}^{K}I(z_{t}=k)\log(q_{t,k})$.
The cross-entropy loss function $\ell_{f_{t}}$ is $\sqrt{2}$-Lipschitz.

Proof. 
When $k=i$, 
\begin{equation*}
    \frac{\partial q_{t,k}}{\partial f_{t,i}}=\frac{\partial \frac{e^{f_{t,i}}}{\sum_{k=1}^{K}e^{f_{t,k}}}}{\partial f_{t,i}}
    =\frac{e^{f_{t,i}}\sum_{k=1}^{K}e^{f_{t,k}}-e^{2f_{t,i}}}{(\sum_{k=1}^{K}e^{f_{t,k}})^{2}}
    =q_{t,i}(1-q_{t,i}).
\end{equation*}

When $k \neq i$, 
\begin{equation*}
    \frac{\partial q_{t,k}}{\partial f_{t,i}} =\frac{\partial \frac{e^{f_{t,k}}}{\sum_{k=1}^{K}e^{f_{t,k}}}}{\partial f_{t,i}}
     =\frac{-e^{f_{t,k}}e^{f_{t,i}}}{(\sum_{k=1}^{K}e^{f_{t,k}})^{2}}
     =-q_{t,k}q_{t,i}.
\end{equation*}
Then,
\begin{equation*}
    \begin{split}
        \frac{\partial \ell_{f_{t}}}{\partial f_{t,i}}&=
        -\sum_{k=1}^{K}I(z_{t}=k)\frac{1}{q_{t,k}}\frac{\partial q_{t,k}}{\partial f_{t,i}}\\
        &=-\frac{I(z_{t}=i)}{q_{t,i}}\frac{\partial q_{t,i}}{\partial f_{t,i}}-\sum_{k \neq i}\frac{I(z_{t}=k)}{q_{t,k}}\frac{\partial q_{t,k}}{\partial f_{t,i}}\\
        &=-\frac{I(z_{t}=i)}{q_{t,i}}q_{t,i}(1-q_{t,i})+\sum_{k \neq i}\frac{I(z_{t}=k)}{q_{t,k}}q_{t,k}q_{t,i}\\
        &=-I(z_{t}=i)+I(z_{t}=i)q_{t,i}+q_{t,i}\sum_{k \neq i}I(z_{t}=k)\\
        &=q_{t,i}-I(z_{t}=i).
    \end{split}
\end{equation*}
\textcolor{black}{
Thus, $\Vert {\partial \ell _{f_{t}}}/{\partial f_{t}}\Vert _{F}^{2}\leq
\sum\nolimits_{k=1}^{K}\left( q_{t,k}^{2}+I(z_{t}=k)\right) =\frac{%
\sum_{k=1}^{K}\left( e^{f_{t,k}}\right)^{2}}{\left(
\sum_{k=1}^{K}e^{f_{t,k}}\right)^{2}}+1\leq 2$. Then, we have  for any \textcolor{black}{$f_{t}, f_{t}^{\prime} \in \mathcal{F}_{t}$}, 
\begin{equation*}
    |\ell_{f_{t}}-\ell_{f_{t}^{\prime}}| \leq \|{\partial \ell_{f_{t}}}/{\partial f_{t}}\|_F
    \|f_{t}-f_{t}^{\prime}\|_F\leq \sqrt{2} \|f_{t}-f_{t}^{\prime}\|_F.
\end{equation*}}
\end{lemma}
Next, we show the Lipschitz continuity of the hinge loss $\ell_{f_{t}}$ in the following lemma.

\begin{lemma}
\label{L11}Let the margin operator be $\psi _{f_{t}}=\max_{k\neq
z_{t}}[q(f_{t})]_{k}-[q(f_{t})]_{z_{t}}$, where $q(f_{t})=q(f_{t}(X_{t}))$.
The hinge loss function $\ell _{f_{t}}=\max \{0,1+\psi _{f_{t}}\}$ is $\sqrt{%
2}$-Lipschitz.

Proof. Let $f_{t}$ and $f_{t}^{\prime }$ be given. Let $k^{\prime }$ satisfy
that $[q(f_{t})]_{k^{\prime }}=\max_{k\neq z_{t}}[q(f_{t})]_{k}$. Then for $%
\psi _{f_{t}}\geq \psi _{f_{t}^{\prime }},$%
\begin{eqnarray}
\psi _{f_{t}}-\psi _{f_{t}^{\prime }} &=&\left( [q(f_{t})]_{k^{\prime
}}-[q(f_{t})]_{z_{t}}\right) -\left( \max_{k\neq z_{t}}[q(f_{t}^{\prime
})]_{k}-[q(f_{t}^{\prime })]_{z_{t}}\right)   \notag \\
&\leq &\left( [q(f_{t})]_{k^{\prime }}-[q(f_{t})]_{z_{t}}\right) -\left(
[q(f_{t}^{\prime })]_{k^{\prime }}-[q(f_{t}^{\prime })]_{z_{t}}\right)
\notag \\
&=&\left( [q(f_{t})]_{k^{\prime }}-[q(f_{t}^{\prime })]_{k^{\prime }}\right)
-\left( [q(f_{t})]_{z_{t}}-[q(f_{t}^{\prime })]_{z_{t}}\right)   \notag \\
&\leq &\sqrt{2}\{\left( [q(f_{t})]_{k^{\prime }}-[q(f_{t}^{\prime
})]_{k^{\prime }}\right) ^{2}+\left( [q(f_{t})]_{z_{t}}-[q(f_{t}^{\prime
})]_{z_{t}}\right) ^{2}\}^{1/2}  \notag \\
&\leq &\sqrt{2}\{\left( [f_{t}]_{k^{\prime }}-[f_{t}^{\prime }]_{k^{\prime
}}\right) ^{2}+\left( [f_{t}]_{z_{t}}-[f_{t}^{\prime }]_{z_{t}}\right)
^{2}\}^{1/2}  \label{EQ:Lip_hinge} \\
&\leq &\sqrt{2}||f_{t}-f_{t}^{\prime }||_{F}.  \notag
\end{eqnarray}%
where (\ref{EQ:Lip_hinge}) holds because the activation function $q$ is $1$%
-Lipschitz. 
Similarly, for $\psi _{f_{t}}\leq \psi _{f_{t}^{\prime }}$, we can show that
$\psi _{f_{t}^{\prime }}-\psi _{f_{t}}\leq \sqrt{2}||f_{t}^{\prime
}-f_{t}||_{F}$. Therefore, $|\psi _{f_{t}}-\psi _{f_{t}^{\prime }}|\leq
\sqrt{2}||f_{t}-f_{t}^{\prime }||_{F}$, so the margin operator $\psi _{f_{t}}
$ is $\sqrt{2}$-Lipschitz. Given that $\ell _{f_{t}}=\max \{0,1+\psi
_{f_{t}}\}$, we have $|\ell _{f_{t}}-\ell _{f_{t}^{\prime }}|\leq $ $|\psi
_{f_{t}}-\psi _{f_{t}^{\prime }}|\leq \sqrt{2}||f_{t}-f_{t}^{\prime }||_{F}$%
, so $\ell _{f_{t}}$ is $\sqrt{2}$-Lipschitz.
\end{lemma}

\section{Proof of Theorem \ref{T3}}\label{S8-2-2}
In this subsection, we calculate the local Rademacher complexity of vanilla RNNs.
Typically, the local Rademacher complexity can be computed by an entropy integral \citep{koltchinskii2000rademacher,bartlett2002localized}, for any $c_{1}>1$ and $\alpha>0$,
\begin{equation*}
    \begin{split}
    \mathfrak{R}_{S}(rB_{F}) &=\mathds{E} \left[\underset{f_{t} \in \widetilde{\mathcal{F}}_{t}, \|f_{t}-f_{t}^{*}\|_{F} \leq r}{\sup}{\frac{1}{n} \sum_{i=1}^{n} \epsilon_{i}f_{t}(X_{it})} \right]\\
    &\leq \underset{\alpha>0}{\inf} \left(\frac{c_{1}}{\sqrt{n}}\int_{\alpha}^{r}\sqrt{\log \mathcal{N}(\mathcal{F}_{t},\epsilon,\|\cdot\|_{F})}\;d\epsilon\right)\\
    &\leq \frac{c_{1}}{\sqrt{n}}\int_{\alpha}^{r}\frac{3\sqrt{3}dg_{t}}{\epsilon}\sqrt{\log(2d^{2})}\;d\epsilon\\
    &\leq \frac{c_{1}}{\sqrt{n}\alpha}3\sqrt{3}rdg_{t}\sqrt{\log(2d^{2})}=\frac{\theta_{t}}{\sqrt{n}}r,
    \end{split}
\end{equation*}
where  $\theta_{t}=\frac{3\sqrt{3}c_{1}}{\alpha}d\sqrt{\log(2d^2)}g_{t}$, $b_{t}=\frac{t(\rho_{h}B_{U})^{t-1}-c_{t}}{\rho_{h}B_{U}-1}$, $c_{t}=\frac{(\rho_{h}B_{U})^{t}-1}{\rho_{h}B_{U}-1}$ and $g_{t}=\rho_{h}B_{x}B_{V}B_{W}\max\{c_{t},\rho_{h}B_{U}b_{t}\}$.
The first inequality follows in \citep{bartlett2002localized} and the second inequality follows from Inequation \ref{L6}. We complete the proof of Theorem \ref{T3}.

\section{Proof of Theorem \ref{T4}}\label{S8-2-3}
In this subsection, we prove the estimation bound of vanilla RNNs. 
We define $\widehat{\mathcal{E}}(f_{t})=\widehat{\mathcal{R}}(f_{t})-\widehat{\mathcal{R}}(f_{t}^{*})$ as the empirical counterpart of the excess risk $\mathcal{E}(f_{t})$  .
We first introduce an event $\Omega_{t}$ on which we can derive the statistical performance of $\widehat{f_{t}}$ using only deterministic arguments: for $\varphi_{t}^{*}$ given in (\ref{E1}) and a positive constant $\theta_{t}$ satisfying ${\rho}A_{t}\theta_{t}<1$,
\begin{equation*}
    \begin{split}
    \Omega_{t} :=\{\text{for\; all}\; f_{t} \in \mathcal{\widetilde{F}}_{t}, |\mathcal{E}(f_{t})-\mathcal{\widehat{E}}(f_{t})|
     \leq {\rho}\theta_{t}\max\left((\varphi_{t}^{*})^{2}, \|f_{t}-f_{t}^{*}\|_{F}^{2}\right)\}.
    \end{split}
\end{equation*}
Let us place ourselves on the event  $\Omega_{t}$. Let $f_{t}$ be a function in $\mathcal{\widetilde{F}}_{t}$ such that $\|f_{t}-f_{t}^{*}\|_{F}>\varphi_{t}^{*}$; then we have
\begin{equation*}
    \begin{split}
  \mathcal{\widehat{E}}(f_{t})&=\left(\mathcal{\widehat{E}}(f_{t})-\mathcal{{E}}(f_{t})\right)+\mathcal{{E}}(f_{t}) \\
  &\geq -{\rho}\theta_{t} \|f_{t}-f_{t}^{*}\|_{F}^{2}+(1/A_{t})\|f_{t}-f_{t}^{*}\|_{F}^{2}\\
 &>(1/A_{t}-{\rho}\theta_{t})(\varphi_{t}^{*})^{2},
  \end{split}
\end{equation*}
where $1/A_{t}-{\rho}\theta_{t}>0$. Since $\mathcal{\widehat{E}}(\widehat{f_{t}}) \leq \Delta_{t}=o\left((\varphi_{t}^{*})^{2}\right)$, we have $\|\widehat{f_{t}}-f_{t}^{*}\|_{F} \leq \varphi_{t}^{*}$ on the event $\Omega_{t}$. Let us now prove the sharp oracle inequality. It follows from
\begin{equation*}
    \begin{split}
        \Delta_{t} & \geq \mathcal{\widehat{E}}(\widehat{f_{t}})=\left(\mathcal{\widehat{E}}(\widehat{f_{t}})-\mathcal{{E}}(\widehat{f_{t}})\right)+\mathcal{{E}}(\widehat{f_{t}}) \\
        &  \geq  -{\rho}\theta_{t}\;\max\left((\varphi_{t}^{*})^{2},\|\widehat{f_{t}}-f_{t}^{*}\|_{F}^{2}\right)+\mathcal{{E}}(\widehat{f_{t}}) \\
        &= -{\rho}\theta_{t}(\varphi_{t}^{*})^{2}+\mathcal{{E}}(\widehat{f_{t}}),
    \end{split}
\end{equation*}
that $\mathcal{E}(\widehat{f_{t}}) \leq {\rho}\theta_{t}(\varphi_{t}^{*})^{2}+\Delta_{t}={\rho}\theta_{t}(\varphi_{t}^{*})^{2}+o\left((\varphi_{t}^{*})^{2}\right)$.

Let us further prove that $\Omega_{t}$ holds at least with the exponential probability from the Theorem \ref{T4}. The proof uses a peeling argument \citep{van2000} of the class $\mathcal{\widetilde{F}}_{t}$ along the shelves
\begin{equation*}
    \mathcal{F}_{t,k}=\{f_{t}\in \mathcal{\widetilde{F}}_{t}: 2^{k-1}\varphi_{t}^{*}<\|f_{t}-f_{t}^{*}\|_{F} \leq 2^{k}\varphi_{t}^{*}\},
\end{equation*}
for all integer $k \geq 1$ and $\mathcal{F}_{t,0}=\mathcal{\widetilde{F}}_{t}\cap(f_{t}^{*}+\varphi_{t}^{*}B_{F})$. The peering argument used here is similar to the one from Proposition 9.2 and 9.3 in \cite{alquier2019}.

For any integer $k$  that by Talagrand's concentration inequality:  for any $\xi>0$, with probability larger than $1-2e^{-\xi}$,
    \begin{equation}\label{E4}  
        Z_{k}\leq 2\mathds{E}Z_{k}+\underset{f_{t}\in \mathcal{F}_{t,k}}{\sup}\sqrt{E\mathcal{L}_{f_{t}}^2}\sqrt{\frac{8\xi}{n}}+\frac{69\|\mathcal{L}_{\mathcal{F}_{t,k}}\|_{\infty}\xi}{2n},
    \end{equation}
where $Z_{k}=\underset{f_{t}\in \mathcal{F}_{t,k}}{\sup} |\mathcal{E}_{f_{t}}-\widehat{\mathcal{E}}_{f_{t}}|$, $\mathcal{L}_{f_{t}}=\ell_{f_{t}}-\ell_{f_{t}^{*}}$,  and $\|\mathcal{L}_{\mathcal{F}_{t,k}}\|_{\infty}=\underset{f_{t}\in \mathcal{F}_{t,k}}{\sup}\|\mathcal{L}_{f_{t}}\|_{\infty}$.

According to Assumption \ref{A5} (ii), one has
$$
\underset{f_{t}\in \mathcal{F}_{t,k}}{\sup}\sqrt{E\mathcal{L}_{f_{t}}^2} \leq \rho\|f_{t}-f_{t}^{*}\|_{F} \leq 2^{k}\rho\varphi_{t}^{*}\;
\text{and}\;\|\mathcal{L}_{\mathcal{F}_{t,k}}\|_{\infty} \leq 2\rho\omega_{t}.
$$
Thus, it only remains to show the upper bound for the expectation $\mathds{E}Z_{k}$. For any function $f_{t}$, we set
\begin{equation*}
    P_{n,\epsilon}f_{t}=\frac{1}{n}\sum_{i=1}^{n}\epsilon_{i}f_{t}(X_{it}).
\end{equation*}
It follows from a symmetrization and a contraction argument (cf. Chapter 4 in \cite{ledoux1991probability}) that
\begin{equation*}
    \begin{split}
    \mathds{E}Z_{k} & \leq 4{\rho}\mathds{E}\;\underset{f_{t}\in \mathcal{F}_{t,k}}{\sup} |P_{N,\epsilon}(f_{t}-f_{t}^{*})|  \leq 4\mathfrak{R}_{S}[(2^{k}\varphi_{t}^{*})B_{F}]\\
     &\leq 4{\rho}\cdot2^{k}\varphi_{t}^{*} \frac{\eta_{t}}{\sqrt{n}}= {\rho}(\theta_{t}/12)2^{k}(\varphi_{t}^{*})^{2},
     \end{split}
\end{equation*}
where the third inequality holds by Theorem \ref{T3} and the last equality holds by (\ref{E1}).

Now, we take $\xi=\nu_{t}\theta_{t}^2n(2^{k-1}\varphi_{t}^{*})^{2}$ in Inequality (\ref{E4}): with probability larger than
\begin{equation*}
    1-2\exp(-\nu_{t}\theta_{t}^2n(2^{k-1}\varphi_{t}^{*})^{2}),
\end{equation*}
for any $f_{t} \in \mathcal{F}_{t,k}$, we have
\begin{equation*}
    \begin{split}
    |\mathcal{E}_{f_{t}}-\widehat{\mathcal{E}}_{f_{t}}|& \leq  {\rho}\frac{\theta_{t}}{3}2^{k-1}(\varphi_{t}^{*})^{2}+4\sqrt{2\nu_{t}}\rho\theta_{t}(2^{k-1}\varphi_{t}^{*})^{2}
    +69\rho\nu_{t}\omega_{t}\theta_{t}^2(2^{k-1}\varphi_{t}^{*})^{2}\\
    & \leq {\rho}\theta_{t} (2^{(k-1)\vee 0}\varphi_{t}^{*})^{2}\left[\frac{1}{3}+4\sqrt{2\nu_{t}}+69\nu_{t}\omega_{t}\theta_{t}\right]\\
    & \leq  {\rho}\theta_{t} \max\left((\varphi_{t}^{*})^{2}, \|f_{t}-f_{t}^{*}\|_{F}^{2}\right),
    \end{split}
\end{equation*}
if $\nu_{t}$ is defined by
\begin{equation*}
    \nu_{t}=\min\left\{\frac{1}{288},\frac{1}{207\theta_{t}\omega_{t}}\right\}.
\end{equation*}
We conclude with a union bound and the probability is larger than \\
$1-2\sum_{k=1}^{\infty} \exp(-\nu_{t}\theta_{t}^{2}n(2^{k-1}\varphi_{t}^{*})^{2})$.
Then, we need to calculate the probability. Let $\vartheta_{t}=n\nu_{t}\theta_{t}^{2}(\varphi_{t}^{*})^{2}$,
\begin{equation*}
    \sum_{k=1}^{\infty} \exp(-\vartheta_{t}4^{k-1}) \geq \sum_{k=1}^{\infty} \exp(-\vartheta_{t}(3k-2)) 
    =\frac{\exp(-\vartheta_{t})}{1-\exp(-3\vartheta_{t})}.
\end{equation*}
 Since $\omega_{t}=o(\sqrt{\eta_{t}})$ holds, $\vartheta_{t} \rightarrow +\infty$.
We complete the proof of Theorem \ref{T4}.

\section{Proof of the Bernstein Condition for the Cross-Entropy Loss}\label{S8-6}


Let $q_{t,k}=\frac{e^{f_{t,k}}}{\sum_{k=1}^{K}e^{f_{t,k}}}$ and $f_{t,k}=f_{t,k}(X_{t})$.
The cross-entropy loss risk  of a function $f_{t}:\mathcal{X}_{t} \rightarrow \mathcal{Y}_{t}$ can be written as
$\mathcal{R}(f_{t})=\mathds{E}[\mu(X_{t},f_{t}(X_{t}))]$ where $\mu(X_{t},f_{t}(X_{t}))=\sum_{k=1}^{K} p_{t,k}\left(\log\sum_{k=1}^{K}e^{f_{t,k}}-f_{t,k}\right)$ and $p_{t,k}=\mathds{P}[z_{t}=k|X_{t}]$ ($k=1,...,K$) is the conditional probability of $z_{t}$ given $X_{t}$.

Since $f_{t}^{*}$ minimizes $f_{t} \rightarrow \mathcal{R}(f_{t})$ over $\mathcal{\widetilde{F}}_{t}$, $\mathds{E}(\nabla \mu[X_{t},f_{t}^{*}(X_{t})][f_{t}(X_{t})-f_{t}^{*}(X_{t})] )\geq 0$. Therefore, it follows from a second-order Taylor expansion that the excess cross-entropy loss for any $f_{t} \in \mathcal{\widetilde{F}}_{t}$ in such that
\begin{equation*}
    \begin{split}
    & \mathcal{E}(f_{t})  =\mathcal{R}(f_{t})-\mathcal{R}(f_{t}^{*})\\
    & \geq \mathds{E}\left\{\frac{1}{2}[f_{t}(X_{t})-f_{t}^{*}(X_{t})]^{T}
     \int_{0}^{1}H[f_{t}^{*}(X_{t})+v(f_{t}(X_{t})-f_{t}^{*}(X_{t}))]\;dv\;[f_{t}(X_{t})-f_{t}^{*}(X_{t})]\right\},
    \end{split}
\end{equation*}
where $H(f_{t}(X_{t}))=\left[\frac{\partial^{2}\mu}{\partial f_{t,i} \partial f_{t,j}}\right]_{K\times K}=
\left[\sum_{k=1}^{K}p_{t,k}\left(I(i=j)q_{t,j}-q_{t,i}q_{t,j}\right)\right]_{K\times K}$. Since $H$ is a positive semi-definite symmetric matrix, a.s. then for every $v \in [0,1]$,
$\mathcal{E}(f_{t}) \geq \frac{\lambda_{\min,t}(H)}{2}\mathds{E}[\|f_{t}(X_{t})-f_{t}^{*}(X_{t})\|_{F}^2]$ and $\lambda_{\min,t}(H)$ is the minimum eigenvalue of a symmetric matrix $H$. We conclude that the cross-entropy loss satisfies Condition \ref{A4} with $A_{t}=\frac{2}{\lambda_{\min,t}(H)}$ with $\lambda_{\min,t}(H)$ satisfying  $\lambda_{\min,t}(H)>0$ .

 \end{appendices}

\section*{CRediT authorship contribution statement}  
\textbf{Xuewei Cheng}: Writing-original draft, Software, Methodology.
\textbf{Ke Huang}: Validation, Software, Investigation.
\textbf{Shujie Ma}: Writing-original draft, Methodology, Writing-review \& editing.

\section*{Declaration of Competing Interest}  
The authors declare that they have no known competing financial interests or personal relationships that could have appeared to influence the work reported in this paper.

\section*{Data availability}  
All datasets used in this work are available online and clearly cited.

\section*{Acknowledgements}
The authors would like to thank the editors and the anonymous reviewers for their valuable comments and suggestions, which led to a substantial improvement of this paper. Ma and Huang's research is supported in part by the U.S. NSF grants DMS-17-12558, DMS-20-14221, and DMS-23-10288. Cheng's research is supported by the Natural Science Foundation of Hunan Province (No.2024JJ6303), the Research Foundation of Education Department of Hunan Province (No.23B0056), and the National Natural Science Foundation of China (No.52101027).

\bibliographystyle{elsarticle-num}
\bibliography{RNN}

\begin{thebibliography}{10}
\expandafter\ifx\csname url\endcsname\relax
  \def\url#1{\texttt{#1}}\fi
\expandafter\ifx\csname urlprefix\endcsname\relax\def\urlprefix{URL }\fi
\expandafter\ifx\csname href\endcsname\relax
  \def\href#1#2{#2} \def\path#1{#1}\fi

\bibitem{sak2014long}
H.~Sak, A.~W. Senior, F.~Beaufays, Long short-term memory recurrent neural network architectures for large scale acoustic modeling, In Fifteenth Annual Conference of the International Speech Communication Association. (2014).

\bibitem{graves2013speech}
A.~Graves, A.-r. Mohamed, G.~Hinton, Speech recognition with deep recurrent neural networks, in: 2013 IEEE international conference on acoustics, speech and signal processing, Ieee, 2013, pp. 6645--6649.

\bibitem{sutskever2014sequence}
I.~Sutskever, O.~Vinyals, Q.~V. Le, Sequence to sequence learning with neural networks, Advances in neural information processing systems 27 (2014).

\bibitem{wu2016google}
Y.~Wu, M.~Schuster, Z.~Chen, Q.~V. Le, M.~Norouzi, W.~Macherey, M.~Krikun, Y.~Cao, Q.~Gao, K.~Macherey, et~al., Google's neural machine translation system: Bridging the gap between human and machine translation, arXiv preprint arXiv:1609.08144 (2016).

\bibitem{xu2015show}
K.~Xu, J.~Ba, R.~Kiros, K.~Cho, A.~Courville, R.~Salakhudinov, R.~Zemel, Y.~Bengio, Show, attend and tell: Neural image caption generation with visual attention, in: International conference on machine learning, PMLR, 2015, pp. 2048--2057.

\bibitem{karpathy2015deep}
A.~Karpathy, L.~Fei-Fei, Deep visual-semantic alignments for generating image descriptions, in: Proceedings of the IEEE conference on computer vision and pattern recognition, 2015, pp. 3128--3137.

\bibitem{cao2018deep}
C.~Cao, F.~Liu, H.~Tan, D.~Song, W.~Shu, W.~Li, Y.~Zhou, X.~Bo, Z.~Xie, Deep learning and its applications in biomedicine, Genomics, proteomics \& bioinformatics 16~(1) (2018) 17--32.

\bibitem{ChenBatsakis2023}
Y.~Chen, S.~Batsakis, V.~H. Poor, Higher-order spatio-temporal neural networks for covid-19 forecasting, in: International Conference on Acoustics, Speech, and Signal Processing 2023, 2023, pp. 1--5.

\bibitem{allen2019convergence}
Z.~Allen-Zhu, Y.~Li, Z.~Song, On the convergence rate of training recurrent neural networks, Advances in neural information processing systems 32 (2019).

\bibitem{vapnik1998statistical}
V.~N. Vapnik, V.~Vapnik, et~al., Statistical learning theory, wiley New York, 1998.

\bibitem{mohri2018}
M.~Mohri, A.~Rostamizadeh, A.~Talwalkar, Foundations of machine learning, MIT press, 2018.

\bibitem{valiant1984theory}
L.~G. Valiant, A theory of the learnable, Communications of the ACM 27~(11) (1984) 1134--1142.

\bibitem{dasgupta1995sample}
B.~Dasgupta, E.~Sontag, Sample complexity for learning recurrent perceptron mappings, Advances in Neural Information Processing Systems 8 (1995).

\bibitem{koiran1998vapnik}
P.~Koiran, E.~D. Sontag, Vapnik-chervonenkis dimension of recurrent neural networks, Discrete Applied Mathematics 86~(1) (1998) 63--79.

\bibitem{Vapnik71}
V.~N. Vapnik, A.~Y. Chervonenkis, On the uniform convergence of relative frequencies of events to their probabilities, Theory of Probability \& Its Applications 16 (1971) 264.

\bibitem{scott2014rademacher}
C.~Scott, Rademacher complexity, Lecture Notes, Statistical Learning Theory 422 (2014).

\bibitem{Huetal2021}
X.~Hu, L.~Chu, J.~Pei, W.~Liu, J.~Bian, Model complexity of deep learning: A survey, Knowledge and Information Systems 63 (2021) 2585–2619.

\bibitem{bartlett2017}
P.~Bartlett, D.~J. Foster, M.~Telgarsky, Spectrally-normalized margin bounds for neural networks, Advances in Neural Information Processing Systems (2017) 6240--6249.

\bibitem{neyshabur2017pac}
B.~Neyshabur, S.~Bhojanapalli, N.~Srebro, A pac-bayesian approach to spectrally-normalized margin bounds for neural networks, arXiv preprint arXiv:1707.09564 (2017).

\bibitem{tu2019understanding}
Z.~Tu, F.~He, D.~Tao, Understanding generalization in recurrent neural networks, in: International Conference on Learning Representations, 2019, pp. 1--17.

\bibitem{zhang2018stabilizing}
J.~Zhang, Q.~Lei, I.~Dhillon, Stabilizing gradients for deep neural networks via efficient svd parameterization, in: International Conference on Machine Learning, PMLR, 2018, pp. 5806--5814.

\bibitem{chen2019}
M.~Chen, X.~Li, T.~Zhao, On generalization bounds of a family of recurrent neural networks, Proceedings of the 23rd International Conference on Artificial Intelligence and Statistics 108 (2020).

\bibitem{vapnik2013nature}
V.~Vapnik, The nature of statistical learning theory, Springer science \& business media, 2013.

\bibitem{vapnik1999nature}
V.~Vapnik, The nature of statistical learning theory, Springer science \& business media, 1999.

\bibitem{koltchinskii2011empirical}
V.~Koltchinskii, Empirical and rademacher processes, in: Oracle Inequalities in Empirical Risk Minimization and Sparse Recovery Problems, Springer, 2011, pp. 17--32.

\bibitem{alquier2019}
P.~Alquier, V.~Cottet, G.~Lecu{\'e}, Estimation bounds and sharp oracle inequalities of regularized procedures with lipschitz loss functions, The Annals of Statistics 47~(4) (2019) 2117--2144.

\bibitem{bartlett2006}
P.~Bartlett, S.~Mendelson, Empirical minimization in Probability Theory and Related Field, Springer, 2006.

\bibitem{Tian2022}
Y.~Tian, D.~Su, S.~Lauria, X.~Liu, Recent advances on loss functions in deep learning for computer vision, Neurocomputing (2022) 129--158.

\bibitem{xie2017all}
D.~Xie, J.~Xiong, S.~Pu, All you need is beyond a good init: Exploring better solution for training extremely deep convolutional neural networks with orthonormality and modulation, in: Proceedings of the IEEE Conference on Computer Vision and Pattern Recognition, 2017, pp. 6176--6185.

\bibitem{fiorillo2021deepsleepnet}
L.~Fiorillo, P.~Favaro, F.~D. Faraci, Deepsleepnet-lite: A simplified automatic sleep stage scoring model with uncertainty estimates, IEEE transactions on neural systems and rehabilitation engineering 29 (2021) 2076--2085.

\bibitem{huang2018new}
X.~Huang, X.~Yang, J.~Zhao, L.~Xiong, Y.~Ye, A new weighting k-means type clustering framework with an l2-norm regularization, Knowledge-Based Systems 151 (2018) 165--179.

\bibitem{liu2021robust}
X.~Liu, H.~Zhao, Robust hierarchical feature selection with a capped l2-norm, Neurocomputing 443 (2021) 131--146.

\bibitem{goodfellow2016deep}
I.~Goodfellow, Y.~Bengio, A.~Courville, Deep learning, MIT press, 2016.

\bibitem{pisier1981remarques}
G.~Pisier, Remarques sur un r{\'e}sultat non publi{\'e} de b. maurey, S{\'e}minaire d'Analyse fonctionnelle (dit" Maurey-Schwartz") (1981) 1--12.

\bibitem{zhang2002covering}
T.~Zhang, Covering number bounds of certain regularized linear function classes, Journal of Machine Learning Research 2~(Mar) (2002) 527--550.

\bibitem{cao2020generalization}
Y.~Cao, Q.~Gu, Generalization error bounds of gradient descent for learning over-parameterized deep relu networks, in: Proceedings of the AAAI Conference on Artificial Intelligence, Vol.~34, 2020, pp. 3349--3356.

\bibitem{vapnik1991principles}
V.~Vapnik, Principles of risk minimization for learning theory, Advances in neural information processing systems 4 (1991).

\bibitem{bartlett2005local}
P.~L. Bartlett, O.~Bousquet, S.~Mendelson, Local rademacher complexities, Annals of Statistics (2005) 1497--1537.

\bibitem{Geoffrey2020}
C.~Geoffrey, L.~Guillaume, L.~Matthieu, Robust high dimensional learning for lipschitz and convex losses, Journal of Machine Learning Research 21 (2020) 1--47.

\bibitem{ledoux1991probability}
M.~Ledoux, M.~Talagrand, Probability in banach spaces, volume 23 of ergebnisse der mathematik und ihrer grenzgebiete (3), Results in Mathematics and Related Areas 3 (1991).

\bibitem{marcinkiewicz1994building}
M.~A. Marcinkiewicz, Building a large annotated corpus of english: The penn treebank, Using Large Corpora 273 (1994).

\bibitem{socher2013recursive}
R.~Socher, A.~Perelygin, J.~Wu, J.~Chuang, C.~D. Manning, A.~Y. Ng, C.~Potts, Recursive deep models for semantic compositionality over a sentiment treebank, in: Proceedings of the 2013 conference on empirical methods in natural language processing, 2013, pp. 1631--1642.

\bibitem{bowman2015large}
S.~Bowman, G.~Angeli, C.~Potts, C.~D. Manning, A large annotated corpus for learning natural language inference, in: Proceedings of the 2015 Conference on Empirical Methods in Natural Language Processing, 2015, pp. 632--642.

\bibitem{hochreiter1997long}
S.~Hochreiter, J.~Schmidhuber, Long short-term memory, Neural computation 9~(8) (1997) 1735--1780.

\bibitem{cho2014learning}
K.~Cho, B.~Van~Merrienboer, C.~Gulcehre, D.~Bahdanau, F.~Bougares, H.~Schwenk, Y.~Bengio, Learning phrase representations using rnn encoder-decoder for statistical machine translation, arXiv preprint arXiv:1406.1078 (2014).

\bibitem{vaswani2017attention}
A.~Vaswani, N.~Shazeer, N.~Parmar, J.~Uszkoreit, L.~Jones, A.~N. Gomez, {\L}.~Kaiser, I.~Polosukhin, Attention is all you need, Advances in neural information processing systems 30 (2017).

\bibitem{beck2024xlstm}
M.~Beck, K.~P{\"o}ppel, M.~Spanring, A.~Auer, O.~Prudnikova, M.~Kopp, G.~Klambauer, J.~Brandstetter, S.~Hochreiter, xlstm: Extended long short-term memory, arXiv preprint arXiv:2405.04517 (2024).

\bibitem{koltchinskii2000rademacher}
V.~Koltchinskii, D.~Panchenko, Rademacher processes and bounding the risk of function learning, in: High dimensional probability II, Springer, 2000, pp. 443--457.

\bibitem{bartlett2002localized}
P.~L. Bartlett, O.~Bousquet, S.~Mendelson, Localized rademacher complexities, in: International Conference on Computational Learning Theory, Springer, 2002, pp. 44--58.

\bibitem{van2000}
S.~A. Van~de Geer, Applications of empirical process theory, volume 6 of Cambridge Series in Statistical and Probabilistic Mathematics, Cambridge University Press, Cambridge, 2000.

\end{thebibliography}

\end{document}